\documentclass[letterpaper, 10 pt, journal, twoside]{IEEEtran}  % Comment this line out if you need a4paper
\pdfoutput=1

\usepackage{graphicx}
\usepackage{bm}
\usepackage{amsmath}
\usepackage{amsfonts}
\usepackage{amssymb}
\usepackage{ascmac}
\usepackage{prettyref}
\usepackage{comment}
\usepackage{multirow}
\usepackage{makeidx}
\usepackage{fancyhdr}
\usepackage{dotseqn}
\usepackage[caption=false,font=footnotesize]{subfig}
\usepackage{float}
\usepackage{cite}
\usepackage{color}
\usepackage{algorithm,algorithmic}
\usepackage{url}

\newrefformat{sec}{\ref{#1}“\nameref{#1}”}
\newrefformat{sub}{\ref{#1}“\nameref{#1}”}
\newrefformat{par}{\ref{#1}“\nameref{#1}”}
\newrefformat{tab}{\ref{#1}“\nameref{#1}”}
\newrefformat{fig}{Fig. \ref{#1}}

\IEEEoverridecommandlockouts                              % This command is only needed if 
\title{\LARGE \bf
Design of non-diagonal stiffness  matrix for assembly task
}

\author{Tsukasa Kusakabe$^{1}$, Sho Sakaino$^{2}$ and Toshiaki Tsuji$^{1}$% <-this % stops a space
\thanks{$^{1}$Tsukasa Kusakabe and Toshiaki Tsuji are with the 
Graduate School of Science and Engineering, Saitama University, 255 Shimo-okubo, Saitama, 338-8570 
{\tt\small E-mail: tsuji@ees.saitama-u.ac.jp}}%
\thanks{$^{2}$Sho Sakaino is with the Graduate School of Systems and
Information Engineering, University of Tsukuba, 1-1-1 Tennodai,
Tsukuba, Ibaraki 305-8577, Japan and the JST PRESTO}%
}
\begin{document}

\maketitle
\thispagestyle{empty}
\pagestyle{empty}

%%%%%%%%%%%%%%%%%%%%%%%%%%%%%%%%%%%%%%%%%%%%%%%%%%%%%%%%%%%%%%%%%%%%%%%%%%%%%%%%
\begin{abstract}
Compliance control is an increasingly employed technique used in the robotic field. 
It is known that various mechanical properties can be reproduced depending on the 
design of the stiffness matrix, but the design theory that takes advantage of this 
high degree of design freedom has not been elucidated.
This paper, therefore, discusses the non-diagonal elements of the stiffness matrix. 
%, which are often ignored in a common setup. 
We proposed a design method according to the 
conditions required for achieving stable motion. 
Additionally, we analyzed the displacement induced by the non-diagonal elements
in response to an external force and found that to   
%and investigate the effects of asymmetrical placement. 
obtain stable contact with a symmetric matrix, the matrix should be positive definite, 
i.e., all eigenvalues must be positive, however its parameter design is complicated. 
In this study, we focused on the use of asymmetric matrices in compliance control  
and showed that the design of eigenvalues can be simplified by using a triangular matrix. 
%Considering that stable control requires the virtual energy function generated by the stiffness control to be convex in the 
%downward direction, we derive an energy function of the asymmetric matrix. Furthermore, we show 
%that setting a triangular matrix when arranging asymmetrically is beneficial to an assembly task. 
This approach expands the range of the stiffness design 
%the range of the non-diagonal elements that can be set in a stable manner 
and enhances the ability of the compliance control to induce motion. We conducted  
experiments using the stiffness matrix and confirmed that assembly could be achieved 
without complicated trajectory planning.
\end{abstract}

%%%%%%%%%%%%%%%%%%%%%%%%%%%%%%%%%%%%%%%%%%%%%%%%%%%%%%%%%%%%%%%%%%%%%%%%%%%%%%%%
\section{INTRODUCTION}
Many in-demand tasks in the field of automation are contact-rich tasks, 
i.e., those involving contact state transitions. In contact rich tasks, the mechanical constraints 
vary considerably with the transition of the contact state between the robot and environment. 
Thus, implementation of the control program becomes complex. 
Various studies have been conducted on contact-rich tasks using the assembly benchmark 
peg-in-hole (PiH) to  expand the scope of automation technology~\cite{review1}. 
In particular, assembly with a clearance of several microns, which is called precision assembly, 
requires accuracy exceeding the positional accuracy of the robot itself, making assembly using 
position control alone difficult. Robots are typically made compliant by control or a similar 
mechanism to conduct assembly without causing collisions due to positional errors. 
The former is called active compliance, and the latter is called passive compliance. 

Impedance and admittance control are widely adopted as control methods that 
form the basis of active compliance~\cite{hogan_imp}. Compliance control approaches generating 
assembly trajectories that satisfy the mechanical constraints of contact can be divided 
into contact-model-based and contact-model-free approaches~\cite{review1}. 
The contact-model-based approach recognizes the contact state from the sensor information 
and calculates the appropriate assembly trajectory based on the contact model~\cite{arie,tang}. 
The contact-model-free approach is a method that uses machine learning. 
Examples include a method called learning from demonstration (LfD) that enables the robot 
to learn from demonstrations given by humans~\cite{LfD1},\cite{IL1}, and a method called learning from 
the environment (LfE) that enables the robot to learn from the data on its state with respect 
to an environment~\cite{RL1,RL2,RL3}. %-\cite{RL3}. 
Furthermore, hand stiffness is considered critical in research focusing on human motion control. 
Thus, to obtain useful information that can be extended to robotic motion control, 
research on estimating the stiffness of the human hand~\cite{human_stiff1} is underway. 
Ajoudani {\it et al.} proposed a tele-impedance control method that uses an estimated 
value of human arm stiffness to control a robot and conduct PiH tasks~\cite{human_stiff2}. 
Variable impedance control for online adaptation to environmental changes has also been 
actively studied~\cite{RL4}.

Methods that use passive compliance obtain the desired stiffness of the hand by using a 
device that combines passive mechanical elements such as springs. The remote center 
compliance (RCC) device is one of the most popular examples, which utilizes the 
inter-axis interference to induce the motion against external force~\cite{RCC}. 
To deal with complicated tasks, variable RCC (VRCC), that can set stiffness in a 
variable manner, is being actively researched~\cite{VRCC1}. 
Lee proposed a VRCC device that operates using a rod 
that adjusts hand stiffness~\cite{VRCC2}. Drigalski {\it et al}. created a compliance hand 
with a wide adaptive error range to overcome the challenge of an error range narrowed 
by mechanical constraints~\cite{Wrist1_wrist}. Furthermore, Hamaya {\it et al}. proposed a 
supplemental method with path planning~\cite{Wrist1_LfE} and LfD~\cite{wrist1_LfD}. 

Development of a mechanical design theory of RCC devices is an important challenge 
to eliminate posture interference due to environmental changes. 
%because 
%an effective design ensures that the translational force generated in the hand does not 
%interfere with the posture of the device. 
Loncaric showed that 
a generic stiffness matrix can be transformed into a normal form in which rotational and translational 
aspects are maximally decoupled by a particular choice of the coordinate frame~\cite{Loncaric}. 
%the generalized stiffness 
%matrix takes a simple form of which translational and rotational stiffness is a diagonal matrix 
%and that it was possible to define a compliance center using 
%this process . 
Roberts expanded on these results by showing that an arbitrary semi-fixed-value 
spatial stiffness matrix could be written in the regular form~\cite{Robert2}. 
%and also stated that any 
%semi-positive-value spatial stiffness matrix could be achieved with a parallel connection 
%of a simple spring and torsion
These design theories aim at eliminating the non-diagonal components of the stiffness 
matrix to suppress inter-axis interference. 
%interference of the force with other axes. 

%The advantage of passive compliance is that 
Although the control 
system can be implemented easily and inexpensively 
%as a result of the mechanical 
%elements absorbing the hand error. However, since the stiffness of the device and 
%the position of the compliance center are designed according to the work content, 
%changing these requires 
using passive compliance, re-design or re-manufacturing 
are often needed for different tasks. 
Since active compliance methods are characterized by the ability to adjust settings 
arbitrarily, applying the RCC principle of inducing motion by adjusting stiffness 
to these methods would be beneficial. % of great benefit.
%To be precise, the passive compliance 
%method is constrained in scope because it is typically a mechanism specialized for a specific task. 
%The stiffness matrix could be arbitrarily operated if this technique was used to implement active compliance, 
%There have been limited number of such attempts. 
Oikawa {\it et al}. focused on the fact that inter-axis interference by non-diagonal elements 
in the stiffness matrix could be used for peg guidance and proposed a method of adjusting 
the stiffness matrix using reinforcement learning (RL)~\cite{oikawa_RAL}.
%The results of this study, which showed that the PiH speed improved 
%with the adjustment of the non-diagonal components, indicate that design of the non-diagonal 
%elements could increase the range of motions that could be generated using variable stiffness 
%control. 
Kozlovsky {\it et al.} showed that the sample inefficiency of on-policy RL improves by reducing 
the action-space and simplifying the policy by learning asymmetric impedance matrices~\cite{kozlovsky}. 
Although some studies show the advantage of the use of non-diagonal elements, the design 
theory of the stiffness matrix remains unestablished. Therefore, in this study, we consider 
the design of the non-diagonal elements of the stiffness matrix, which are key to achieving motion 
guidance. Particularly, we focus on the fact that compliance control can use asymmetric matrices 
and show that the design of eigenvalues can be simplified by using a triangular matrix. 

This approach expands the range of the non-diagonal elements of the stiffness matrix and 
enhances the ability of the compliance control to induce motion. We conduct  
experiments using the stiffness matrix and confirm that assembly can be achieved 
without complicated trajectory planning.

\if0
\begin{figure}[tb]
    \centering
        \includegraphics[width=84mm]{./fig/Contact_state_transition_figure.pdf}
        \caption{Transition flow after parts contact}
        \label{fig:contact_state_transition_figure}
\end{figure}

In particular, it differs from other studies in that it designs a stiffness matrix suitable for 
the generated trajectory in step 3). 
As many model-based PiH studies, it is also possible to complete the contact-rich tasks by 
only planning a trajectory depending on the contact state. But it has been shown that 
the induction using compliance control has an effect of improving the response of contact 
motion~\cite{oikawa_ellipse}.    
%Literature have shown that environmental adaptive behavior can be 
%acquired by generating trajectories according to the state[*][*], and it is also well known that
%compliance control has the effect of improving that environmental adaptability[*][*].
Impedance control researches have shown that generating axes with higher compliance can
guide the motion and facilitate execution of contact-rich tasks~\cite{balatti2020}. 
The contribution of this study is to clarify the design method of the stiffness matrix  
for the induction against external force during contact rich tasks.
\fi

\section{Motion induction based on compliance control}
We propose a method to design a control system using the following steps.
\begin{enumerate}
\item Specify the contact conditions that can occur in contact-rich tasks.
\item Design a path from all possible contact states to the goal contact state.
\item Generate a trajectory to transition to the next contact state toward the goal, and design a stiffness matrix suitable for that trajectory.
\end{enumerate}
%In the proposed method, a guideline is provided for designing a stiffness matrix that is effective 
%for peg-in-hole (PiH). %, which is a benchmark task in assembly. 
In this section, we first describe the use of non-diagonal terms of compliance control 
to induce motion, and then we explain basics of admittance control. 
After describing the design of a non-diagonal matrix assuming a symmetric matrix, we extend the 
method to the asymmetric case. 
We then discuss the advantage of a triangular matrix, and finally the 
implementation method for an actual PiH is described.
\subsection{Path planning considering different contact states}
\begin{figure}[tb]
    \centering
        \includegraphics[clip,width=8.0cm]{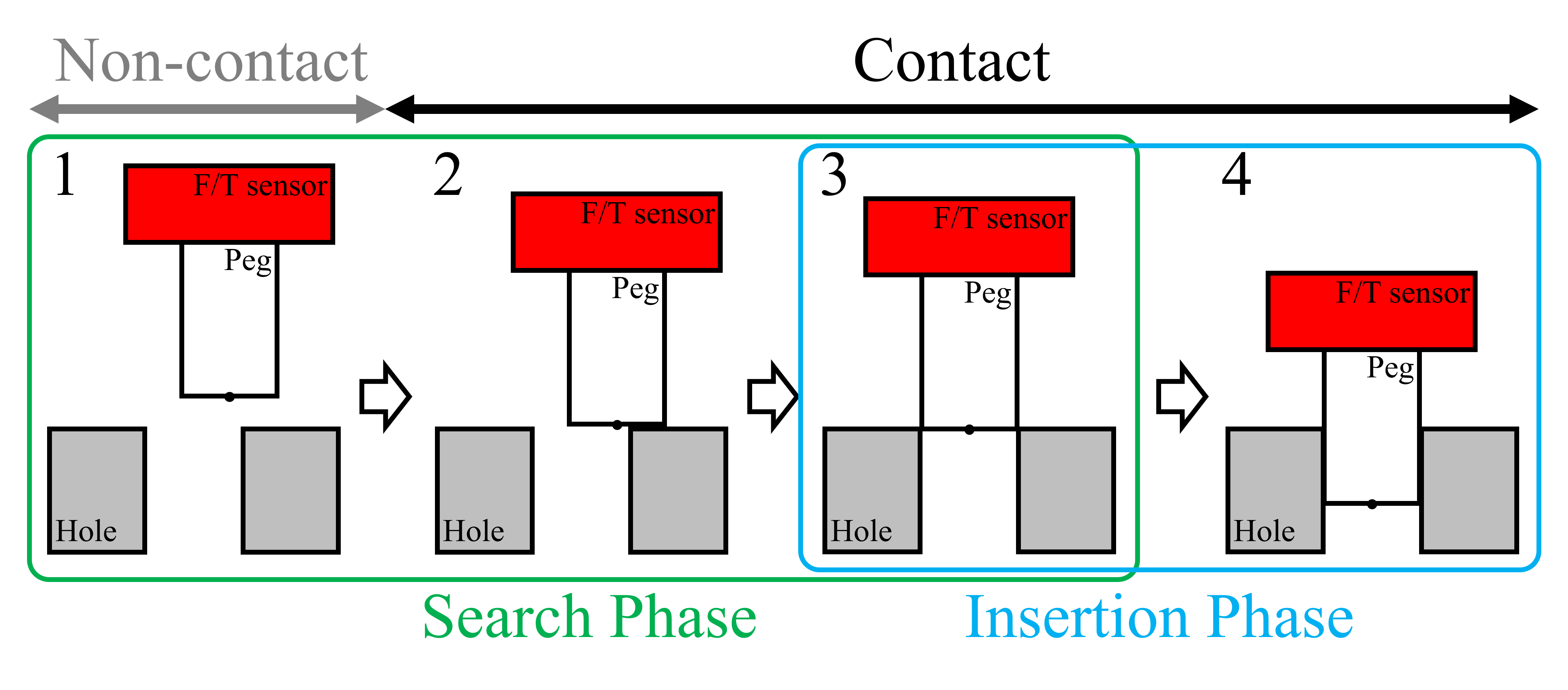}
        \caption{Phases of peg-in-hole}
        \label{peg-in-hole_phase}
\end{figure}

\begin{figure}[tb]
    \centering
        \includegraphics[clip,width=8.0cm]{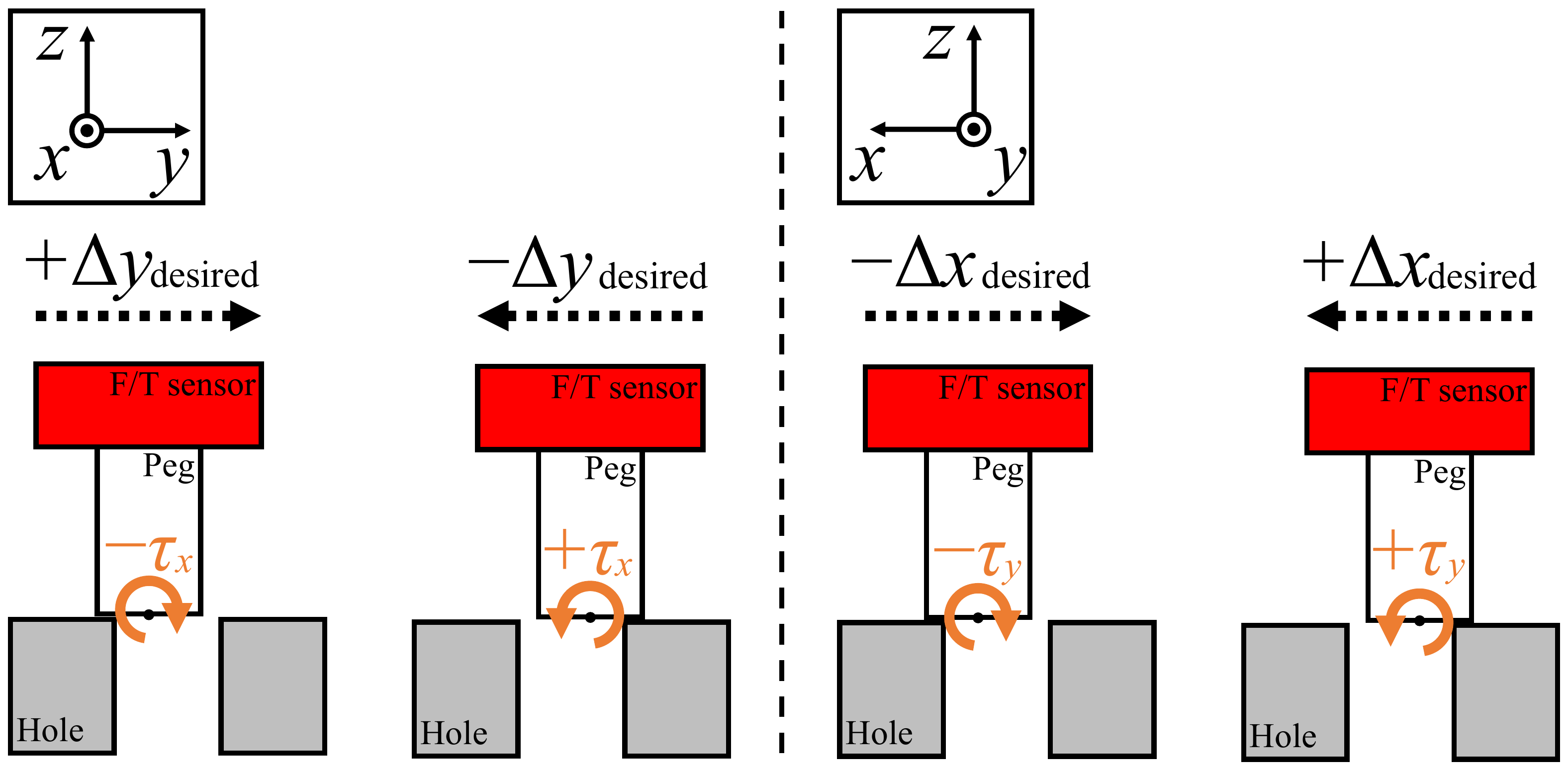}
        \caption{Search phase of peg-in-hole}
        \label{peg-in-hole_saguri}
\end{figure}

%As shown in Fig. \ref{peg-in-hole_phase}4.1, 
%PiH progresses from a Search phase into an Insertion phase. However, the Search phase is a challenging 
%component of the PiH task. This is because errors such as environmental errors, robot hand position 
%errors, and sensor errors are typically generated during this phase. 
Fig. \ref{peg-in-hole_phase} shows phases of the PiH in this study. 
For simplicity, an example is shown where both the cylindrical hole 
and the peg are perpendicular to the plate. 
After the peg contacts the surface, the phases can be divided into the search and insertion phase. 
Fig. \ref{peg-in-hole_saguri} shows the force and moment in the search phase at the moment of 
contact with the environment. 
%Here, in this method, 
%we focus on the case where a position error occurs.% , and posture errors are not considered. 
From Fig.  \ref{peg-in-hole_saguri}, it can be seen that the sign of each moment $\tau_y$ and $\tau_x$ applied to the hand 
corresponds to the direction of the hole when an error is generated in each x and y direction, 
respectively. After an error has been generated, it is possible to guide the peg toward the hole by generating 
position displacement in response to the moment. 
Therefore, we consider the design of non-diagonal elements 
that induce the desired positional displacements $\Delta x_{desired}$ and 
$\Delta y_{desired}$ by $\tau_y$ and $\tau_x$, respectively, and extend it to the asymmetrically 
arranged stiffness matrix. 
%Accordingly, the non-diagonal elements to be designed are $k_{xry}$ and $k_{yrx}$. 
%In the proposed method, a stiffness matrix that compensates 
%for the assumed position error is designed by setting the desired 
%displacement amount to $\Delta$ and using it to substitute the maximum 
%position error that could occur.
%due to the environment, robot, sensor, etc.


\subsection{Admittance control and its stiffness matrix}
We adopt admittance control based on (\ref{admmitance_formula}) for compliance control. 
%Admittance control is a form of compliance control, in which the 
%manipulator’s hand is set as a virtual admittance model, and its behavior 
%is satisfied. By setting the impedance of the robot hand as mass $\bm{M}$, 
%viscosity $\bm{D}$, and stiffness $\bm{K}$, the admittance control involving 
%the control of the hand satisfies 
%Eq. .
\begin{eqnarray}
    \bm{F}_{\mathrm{res}} = \bm{M}\Delta\bm{\ddot{X}}^{\mathrm{adm}}_{\mathrm{cmd}} + \bm{D}\Delta\bm{\dot{X}}^{\mathrm{adm}}_{\mathrm{cmd}} + \bm{K}\Delta\bm{X}^{\mathrm{adm}}_{\mathrm{cmd}}
    \label{admmitance_formula}
\end{eqnarray}

\begin{eqnarray}
    \bm{f}
    =
    \begin{bmatrix}
        f_{x} \\
        f_{y} \\
        f_{z} \\
    \end{bmatrix},~
    \bm{\tau}
    =
    \begin{bmatrix}
    \tau_{x} \\
    \tau_{y} \\
    \tau_{z} \\
    \end{bmatrix},~
    \bm{F}_{\mathrm{res}}
    =
    \begin{bmatrix}
    \bm{f} \\
    \bm{\tau} 
    \end{bmatrix} \nonumber
    \label{admmitance_formula2}
\end{eqnarray}
\begin{eqnarray}
    \Delta \bm{x}
    =
    \begin{bmatrix}
        \Delta x \\
        \Delta y \\
        \Delta z \\
    \end{bmatrix},~
    \Delta \bm{r}
    =
    \begin{bmatrix}
        \Delta r_{x} \\
        \Delta r_{y} \\
        \Delta r_{z} \\
    \end{bmatrix},~
    \Delta \bm{X}^{\mathrm{adm}}_{\mathrm{cmd}}
    =
    \begin{bmatrix}
        \Delta \bm{x} \\
        \Delta \bm{r} 
    \end{bmatrix} \nonumber
    \label{admmitance_formula3}
\end{eqnarray}
Here,  $\bm{M}$, $\bm{D}$ and $\bm{K}$ denote 
mass, viscosity, and stiffness of the admittance model, respectively. 
$f$, $m$, and $\Delta r$ denote force, moment, postural displacement, respectively. 
$\Delta x$, $\Delta y$ and $\Delta z$ denote positional displacement in the x, y, and z axis, respectively. 
%When the robot comes into contact with the environment, $\bm{F}_{\mathrm{res}}$ 
%is generated by the impedance values of the robot and environment. Admittance 
%control outputs  with $\bm{F}_{\mathrm{res}}$ as the input 
%so as to satisfy Eq. (\ref{admmitance_formula}). 

\begin{figure}[tb]
    \centering
        \includegraphics[width=80mm]{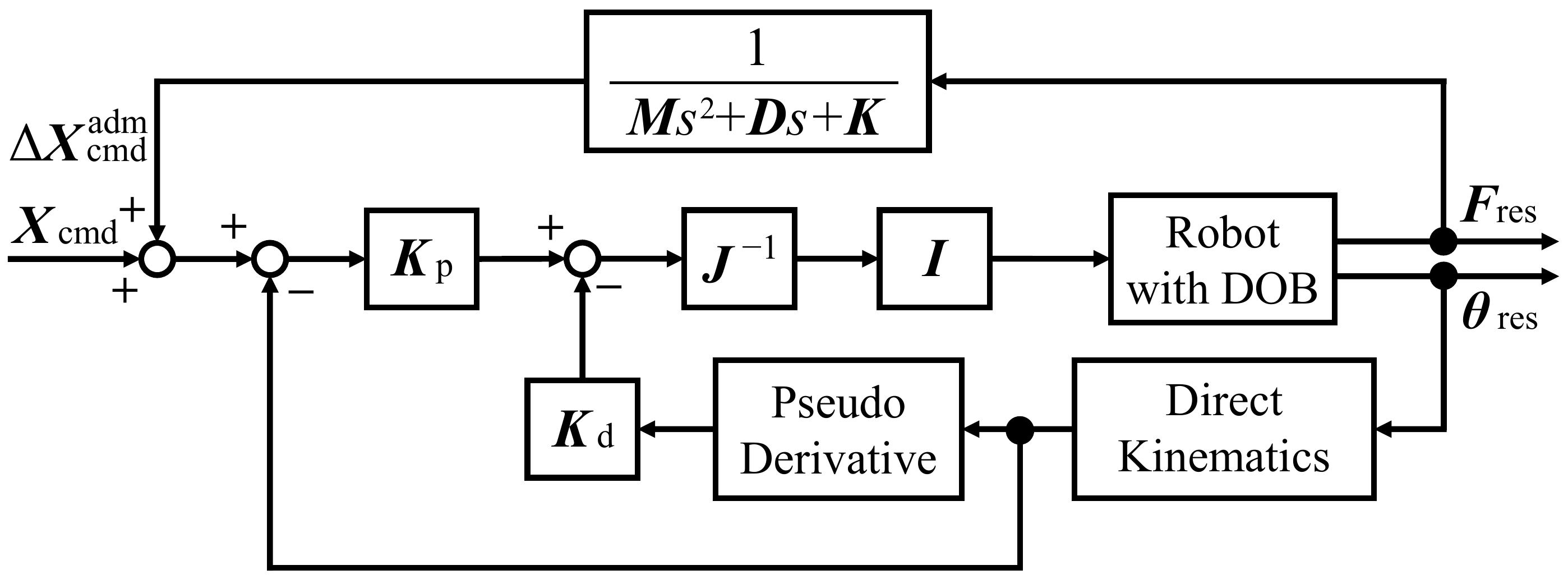}
        \caption{Blockdiagram of admittance control.}
        \label{fig:blockdiagram}
\end{figure}

Fig. \ref{fig:blockdiagram} shows a block diagram of the control system.
The stiffness matrix, $\bm{K}$, represents the static ratio of the displacement $\bm{\Delta x}$ 
and force/moment $\bm{F}$.  $\bm{K}_p$ and $\bm{K}_v$ denote the proportional and derivative gain, 
respectively. $\bm{J}$ and $\bm{I}$ denote the Jacobian and inertia matrix, respectively.  
$\bm{\theta}_{res}$ denotes the joint angle response and $\bm{X}_{cmd}$ denotes the position command. 
The methods described below are also applicable to impedance control systems.
%displacement and postural displacement, $\Delta \bm{X}=[\Delta \bm{x}, \Delta \bm{r}]^{\mathrm T}$, 
%in an arbitrary direction according to the available 
%information on an arbitrary force/moment, $\bm{F}=[\bm{f}, \bm{m}]^{\mathrm T}$, 
%applied to the robot’s hand. These relationships can be expressed as 
%Eqs. (\ref{stiffness_matrix})  and (\ref{stiffness_matrix2}).
\begin{eqnarray}
    \bm{F} &=& \bm{K} \Delta \bm{X} \label{stiffness_matrix} \\
    \!\!\!\!\!\!\!\!\!\!\!\!\!\!\!\!\!\!\!\!\!\!\!\!
    \begin{bmatrix}
        f_{x} \\
        f_{y} \\
        f_{z} \\
        \tau_{x} \\
        \tau_{y} \\
        \tau_{z} \\
    \end{bmatrix}
    \!\!\!\!\!\!&=&\!\!\!\!\!\!
    \begin{bmatrix}
        k_{xx} \!\!& k_{xy} \!\!& k_{xz} \!\!& k_{xr_x} \!\!& k_{xr_y} \!\!& k_{xr_z} \\
        k_{yx} \!\!& k_{yy} \!\!& k_{yz} \!\!& k_{yr_x} \!\!& k_{yr_y} \!\!& k_{yr_z} \\
        k_{zx} \!\!& k_{zy} \!\!& k_{zz} \!\!& k_{zr_x} \!\!& k_{zr_y} \!\!& k_{zr_z} \\
        k_{r_xx} \!\!& k_{r_xy} \!\!& k_{r_xz} \!\!& k_{r_xr_x} \!\!& k_{r_xr_y} \!\!& k_{r_xr_z} \\
        k_{r_yx} \!\!& k_{r_yy} \!\!& k_{r_yz} \!\!& k_{r_yr_x} \!\!& k_{r_yr_y} \!\!& k_{r_yr_z} \\
        k_{r_zx} \!\!& k_{r_zy} \!\!& k_{r_zz} \!\!& k_{r_zr_x} \!\!& k_{r_zr_y} \!\!& k_{r_zr_z} \\
    \end{bmatrix}\!\!\!\!
    \begin{bmatrix}
        \Delta x \\
        \Delta y \\
        \Delta z \\
        \Delta r_{x} \\
        \Delta r_{y} \\
        \Delta r_{z} \\
    \end{bmatrix} \label{stiffness_matrix2}
\end{eqnarray}
%Eq. (\ref{stiffness_matrix2}) shows the relationship between the force/moment 
%applied to the robot’s hand by $\bm{K}$ and the positional displacement/postural 
%displacement. 
Here, using the inverse matrix, $\bm{K^{-1}}$, the relationship is established as:   
\begin{eqnarray}
    \Delta \bm{X} &=& \bm{K}^{\mathrm{-1}}\bm{F} \label{stiffness_matrix_inverse} \\
    \!\!\!\!\!\!\!\!\!\!\!\!\!\!\!\!\!\!\!\!\!\!\!\!
    \begin{bmatrix}
        \Delta x \\
        \Delta y \\
        \Delta z \\
        \Delta r_{x} \\
        \Delta r_{y} \\
        \Delta r_{z} \\
    \end{bmatrix}
    \!\!\!\!\!\!&=&\!\!\!\!\!\!
    \begin{bmatrix}
        k^{\mathrm{inv}}_{xx} \!\!& k^{\mathrm{inv}}_{xy} \!\!& k^{\mathrm{inv}}_{xz} \!\!& k^{\mathrm{inv}}_{xr_x} \!\!& k^{\mathrm{inv}}_{xr_y} \!\!& k^{\mathrm{inv}}_{xr_z} \\
        k^{\mathrm{inv}}_{yx} \!\!& k^{\mathrm{inv}}_{yy} \!\!& k^{\mathrm{inv}}_{yz} \!\!& k^{\mathrm{inv}}_{yr_x} \!\!& k^{\mathrm{inv}}_{yr_y} \!\!& k^{\mathrm{inv}}_{yr_z} \\
        k^{\mathrm{inv}}_{zx} \!\!& k^{\mathrm{inv}}_{zy} \!\!& k^{\mathrm{inv}}_{zz} \!\!& k^{\mathrm{inv}}_{zr_x} \!\!& k^{\mathrm{inv}}_{zr_y} \!\!& k^{\mathrm{inv}}_{zr_z} \\
        k^{\mathrm{inv}}_{r_xx} \!\!& k^{\mathrm{inv}}_{r_xy} \!\!& k^{\mathrm{inv}}_{r_xz} \!\!& k^{\mathrm{inv}}_{r_xr_x} \!\!& k^{\mathrm{inv}}_{r_xr_y} \!\!& k^{\mathrm{inv}}_{r_xr_z} \\
        k^{\mathrm{inv}}_{r_yx} \!\!& k^{\mathrm{inv}}_{r_yy} \!\!& k^{\mathrm{inv}}_{r_yz} \!\!& k^{\mathrm{inv}}_{r_yr_x} \!\!& k^{\mathrm{inv}}_{r_yr_y} \!\!& k^{\mathrm{inv}}_{r_yr_z} \\
        k^{\mathrm{inv}}_{r_zx} \!\!& k^{\mathrm{inv}}_{r_zy} \!\!& k^{\mathrm{inv}}_{r_zz} \!\!& k^{\mathrm{inv}}_{r_zr_x} \!\!& k^{\mathrm{inv}}_{r_zr_y} \!\!& k^{\mathrm{inv}}_{r_zr_z} \\
    \end{bmatrix}\!\!\!\!
    \begin{bmatrix}
        f_{x} \\
        f_{y} \\
        f_{z} \\
        \tau_{x} \\
        \tau_{y} \\
        \tau_{z} \\
    \end{bmatrix}
    \label{stiffness_matrix_inverse2}
\end{eqnarray}

Compliance control achieves a compliance behavior for a tip defined at a single point on the end-effector. 
While the external force acting on the tip can be replaced by the external force of the sensor, 
there is a difference between the moment acting on the tip and the moment detected by the sensor 
in proportion to the distance between them, as shown in (\ref{eq:intrinsic}). 
\begin{eqnarray}
    \bm{f}^{\mathrm {sensor}} &=&  \bm{f}^{\mathrm {peg}}\label{eq:force}\\
    \bm{\tau}^{\mathrm {sensor}} &=&  \bm{\tau}^{\mathrm {peg}} + \bm{l}^{\mathrm {e}}\bm{f}^{\mathrm {peg}}\label{eq:intrinsic}
\end{eqnarray}
where the superscripts sensor and peg denote the sensor response and actual value on the tip of the peg, respectively. 
$\bm{l}^{\mathrm {e}}$ denotes the relateive position from the sensor to the tip of the peg. 
Therefore, the moment on the tip should be calculated by the following formula: 
\begin{eqnarray}
    \bm{\tau}^{\mathrm {peg}} &=&  \bm{\tau}^{\mathrm {sensor}} - \bm{l}^{\mathrm {e}}\bm{f}^{\mathrm {sensor}}\label{m_peg}
\end{eqnarray}
If any external force acts on the tip, the formula cancels this error, but in practice the contact point 
moves depending on the motion of the robot. Therefore, in any compliance control method, 
the external force is considered to be acting on the tip, and the gain design is based on the 
assumption that errors due to fluctuations in the contact position will occur. 
In the PiH, the tip is defined as the center point of the bottom of the peg, 
since external forces are expected to mainly act on the bottom of the peg. 

\subsection{Design of symmetric stiffness matrix}
In this subsection, we describe the properties of the diagonal and non-diagonal elements of the 
stiffness matrix and its inverse. 
%, taking the diagonal and non-diagonal elements of $\bm{K}^{-1}$ as examples.
First, a diagonal element $k^{inv}_{yy}$ 
has the effect of inducing displacement $\Delta y$ in the translational direction 
according to the force $f_y$ applied in the translational direction. In other words, the 
diagonal elements of the stiffness matrix induce positional/postural displacement along 
the same axis as the applied force/moment. Similarly, non-diagonal elements $k^{inv}_{yr_x}$ of the 
stiffness matrix have the effect of inducing displacement $\Delta y$ in the translational direction 
according to moment $\tau_x$ applied to the rotational direction. Thus, all the non-diagonal 
elements of the stiffness matrix induce displacement along an axis that is different from 
the direction of the applied force/moment. The non-diagonal element values represent the 
amount of interference, and their arrangement determines which axis causes the interference. 
Although inter-axis interference can be manipulated by adjusting the values of each element, 
the system can be unstable depending on the parameter settings, unlike mechanical passive mechanisms. 
Therefore, focusing on the eigenvalues of the stiffness matrix, we designed the parameters so that 
divergent vibration modes do not occur.
%its effect on the stability of the control system must be carefully considered.

%When arranging the non-diagonal elements of the stiffness matrix symmetrically, the control 
%system might become unstable. 
%This depends on the set value; however, 
Based on an energy function derived from a symmetric stiffness matrix, passivity of the 
admittance model can be ensured 
by setting the stiffness matrix to a definite-value symmetric matrix, as in the case of 
mechanical elastic structures~\cite{kusakabe}. For this to be the case, all eigenvalues $\lambda$ of a matrix must 
be positive. For example, if the case of a two-dimensional stiffness matrix is expressed as

\begin{eqnarray}
    \begin{bmatrix}
        f_{y} \\
        f_{z} \\
    \end{bmatrix}
    =
    \begin{bmatrix}
        k_{yy} & k_{yz} \\
        k_{zy} & k_{zz} \\
    \end{bmatrix}
    \begin{bmatrix}
        \Delta{y} \\
        \Delta{z} \\
    \end{bmatrix}
    =
    \begin{bmatrix}
        k_{yy} & k_{\mathrm{n}} \\
        k_{\mathrm{n}} & k_{zz} \\
    \end{bmatrix}
    \begin{bmatrix}
        \Delta{y} \\
       \Delta{z} \\
    \end{bmatrix}
    \label{stiffness_matrix_yz_sym}
\end{eqnarray}
\begin{eqnarray}
    \begin{bmatrix}
        \Delta{y} \\
        \Delta{z} \\
    \end{bmatrix}
    =
    \begin{bmatrix}
        k^{\mathrm{inv}}_{yy} & k^{\mathrm{inv}}_{yz} \\
        k^{\mathrm{inv}}_{zy} & k^{\mathrm{inv}}_{zz} \\
    \end{bmatrix}
    \begin{bmatrix}
        f_{y} \\
        f_{z} \\
    \end{bmatrix}
    =
    \begin{bmatrix}
        k^{\mathrm{inv}}_{yy} & k^{\mathrm{inv}}_{\mathrm{n}} \\
        k^{\mathrm{inv}}_{\mathrm{n}} & k^{\mathrm{inv}}_{zz} \\
    \end{bmatrix}
    \begin{bmatrix}
        f_{y} \\
        f_{z} \\
    \end{bmatrix}
    \label{stiffness_matrix_inverse_yz_sym}
\end{eqnarray}
then its eigen equation is
\begin{eqnarray}
    % \mathrm{det}(\bm{K} - \lambda \bm{E}) &=& 0 \label{lambda1}\\
    %                              &=&(k_{yy}-\lambda)(k_{zz}-\lambda)-k_{\mathrm{n}}^2=0 \label{lambda2}
    (k_{yy}-\lambda)(k_{zz}-\lambda)-k_{\mathrm{n}}^2=0. \label{lambda2}
\end{eqnarray}
Subsequently, the condition where the eigenvalues $\lambda$ become positive can be determined as follows:
\begin{eqnarray}
    k_{yy}k_{zz}>k_{n}^2  \label{lambda6} 
\end{eqnarray}
It is possible in principle to extend this to 6 dimensions and design parameters 
so that all eigenvalues are positive. However, as the number of parameters 
increases, the conditionals become more complex.

\subsection{Asymmetrically arranged stiffness matrix design}\label{Asymmetric matrix}
%Before considering the case of arranging the non-diagonal elements of the stiffness matrix asymmetrically, 
%the case of symmetrically arranged non-diagonal elements must be re-assessed. 
As mentioned in the previous 
subsection, cases of symmetrical arrangements must satisfy the conditions of positive-definite symmetry. 
This indicates that when the stiffness matrix (which is a real symmetric matrix) is diagonalized, the eigenvalues 
of the stiffness matrix must be positive. However, it also means that the elastic energy generated by 
the displacement must always be positive. Moreover, the diagonalized matrix can be considered as the stiffness 
on each co-axis. Thus, the stiffness matrix with symmetrically arranged non-diagonal elements represents the 
distribution where the stiffness in each axis is rotated arbitrarily, known as the stiffness ellipse.
Even in the case of asymmetrically arranged non-diagonal elements, the eigenvalues of the stiffness matrix, 
which are related to the square of the modal frequencies~\cite{doebling},   
must be positive. 
%in the designed virtual stiffness matrix becase they present the 
%time constant of vibration modes. 
%This ensures a positional displacement output 
%that is stable with respect to the force. To be precise, the design must ensure that the energy of the asymmetric 
%stiffness matrix comprising the non-diagonal elements must be positive. 
Here, as an example, we determine the 
conditions to be satisfied by asymmetric stiffness matrix $\bm{K}$ for a two-dimensional space (yz plane).
\begin{eqnarray}
    \bm{K}
    =
    \begin{bmatrix}
        k_{yy} & k_{yz} \\
        k_{zy} & k_{zz} \\
    \end{bmatrix}
    =
    \begin{bmatrix}
        k_{yy} & k_{yz} \\
        0 & k_{zz} \\
    \end{bmatrix}
    \label{stiffness_matrix_yz_asym}
\end{eqnarray}
is set, and the relationship with the external force F and the positional displacement $\Delta \bm{X}$ is expressed as follows:
\begin{eqnarray}
    \begin{bmatrix}
        f_{y} \\
        f_{z} \\
    \end{bmatrix}
    =
    \begin{bmatrix}
        k_{yy} & k_{yz} \\
        0 & k_{zz} \\
    \end{bmatrix}
    \begin{bmatrix}
        \Delta{y} \\
        \Delta{z} \\
    \end{bmatrix}
    \label{stiffness_matrix_yz_asym2}
\end{eqnarray}
\begin{eqnarray}
    \begin{bmatrix}
        \Delta{y} \\
        \Delta{z} \\
    \end{bmatrix}
    =
    \begin{bmatrix}
        k^{\mathrm{inv}}_{yy} & k^{\mathrm{inv}}_{yz} \\
        0 & k^{\mathrm{inv}}_{zz} \\
    \end{bmatrix}
    \begin{bmatrix}
        f_{y} \\
        f_{z} \\
    \end{bmatrix}
    \label{stiffness_matrix_yz_asym3}
\end{eqnarray}
From (\ref{stiffness_matrix_yz_asym2}) and (\ref{stiffness_matrix_yz_asym3}), 
the external force $f_y$ only changes displacement $\Delta y$ in the y-direction, 
whereas external force $f_z$ induces a displacement in both $\Delta y$ and 
$\Delta z$. Thus, when arranging the stiffness matrix asymmetrically in a 
triangular matrix, only specific axes will interfere unilaterally.
%Here, we consider the elastic energy of the asymmetrically arranged 
%spring distribution. As shown in (\ref{stiffness_matrix}), stiffness matrix $\bm{K}$ 
%follows Hooke’s law with respect to the position and force. Hence, the elastic 
%energy is expressed as follows:

The conditions where the eigenvalues become positive are as follows: 
%conditions for the stiffness matrix are given by:
\begin{eqnarray}
    k_{yy}>0 \wedge k_{zz}>0. \label{condition} 
\end{eqnarray}
It is known that the eigenvalues of a triangular matrix coincide with 
the diagonal components of the matrix. Therefore, when (\ref{condition}) 
is extended to 6 dimensions, the conditions are obtained as follows: 
\begin{eqnarray}
    k_{xx}>0 &\!\!\!\!\wedge&\!\!\!\! k_{yy}>0 \wedge k_{zz}>0 \wedge \nonumber\\
   &&\!\!\!\!\!\!\!\!k_{r_xr_x}>0 \wedge k_{r_yr_y}>0 \wedge k_{r_zr_z}>0. \label{condition6} 
\end{eqnarray}
There are no theoretical constraints on the non-diagonal components because they 
do not affect the eigenvalues. 

It should be noted here that the elastic energy of the asymmetrical stiffness matrix 
cannot be determined in the same manner as that for a symmetrical stiffness matrix. 
This is because an asymmetrical stiffness matrix is not a positive-definite symmetric 
matrix and it does not satisfy the Maxwell-Betti reciprocal work theorem, which states 
that the spring arrangement must cause equivalent interference on the axes. 
\if 0
Here, work values Wfy and Wfz are expressed as follows:
$\cdots$ (45)(46)\\
\begin{eqnarray}
    W_{f_y} &=& f_y \Delta y \label{work1} \\
    W_{f_z} &=& f_z \Delta y + f_z \Delta z = f_z \Delta z \label{work2}
\end{eqnarray}
and $\Delta y= \Delta zcos2= 0$. Arbitrary stiffness control of the robot’s 
hand requires the prevention of the virtual energy generated by stiffness
 control from discharging so that the entire control system does not become unstable.
The elastic energy in each axis due to stiffness matrix $\bm{K}$ is as follows:
\begin{eqnarray}
    U_{f_y} &=& \frac{1}{2} k_{yy} \Delta y^2 \label{work3} \\
    U_{f_z} &=& \frac{1}{2} k_{zz} \Delta z^2 \label{work4}
\end{eqnarray}
Therefore, from Eqs. (\ref{work1}) and (\ref{work2}), 
\fi

%The characteristics of the spring show that the diagonal elements of the 
%stiffness matrix are positive because the potential energy for the expansion 
%and contraction of the spring at the reference value becomes negative. 
%In particular, when stiffness matrix $\bm{K}$ is an upper or lower triangular 
%matrix, the eigenvalues of the stiffness matrix will always be positive 
%provided that (\ref{condition}) is satisfied, even if each non-diagonal 
%element has an arbitrary real value. This is because eigenvalues of stiffness 
%matrix K that satisfies an upper or lower triangular matrix will each become 
%diagonal elements and will always be positive, provided that the diagonal 
%elements of stiffness matrix $\bm{K}$ are positive. 
Furthermore, the 
determinant $\mathrm{det}(\bm{K})$ is always positive when the matrix is 
inverted, which shows the ease of design. 
Thus, the non-diagonal elements 
have a larger range of values compared with the symmetrical matrices and the 
configurable range of the stiffness matrix expands.

\subsection{Task-specific design of stiffness matrix}
\begin{figure}[t]
        \begin{center}
            \includegraphics[width=60mm]{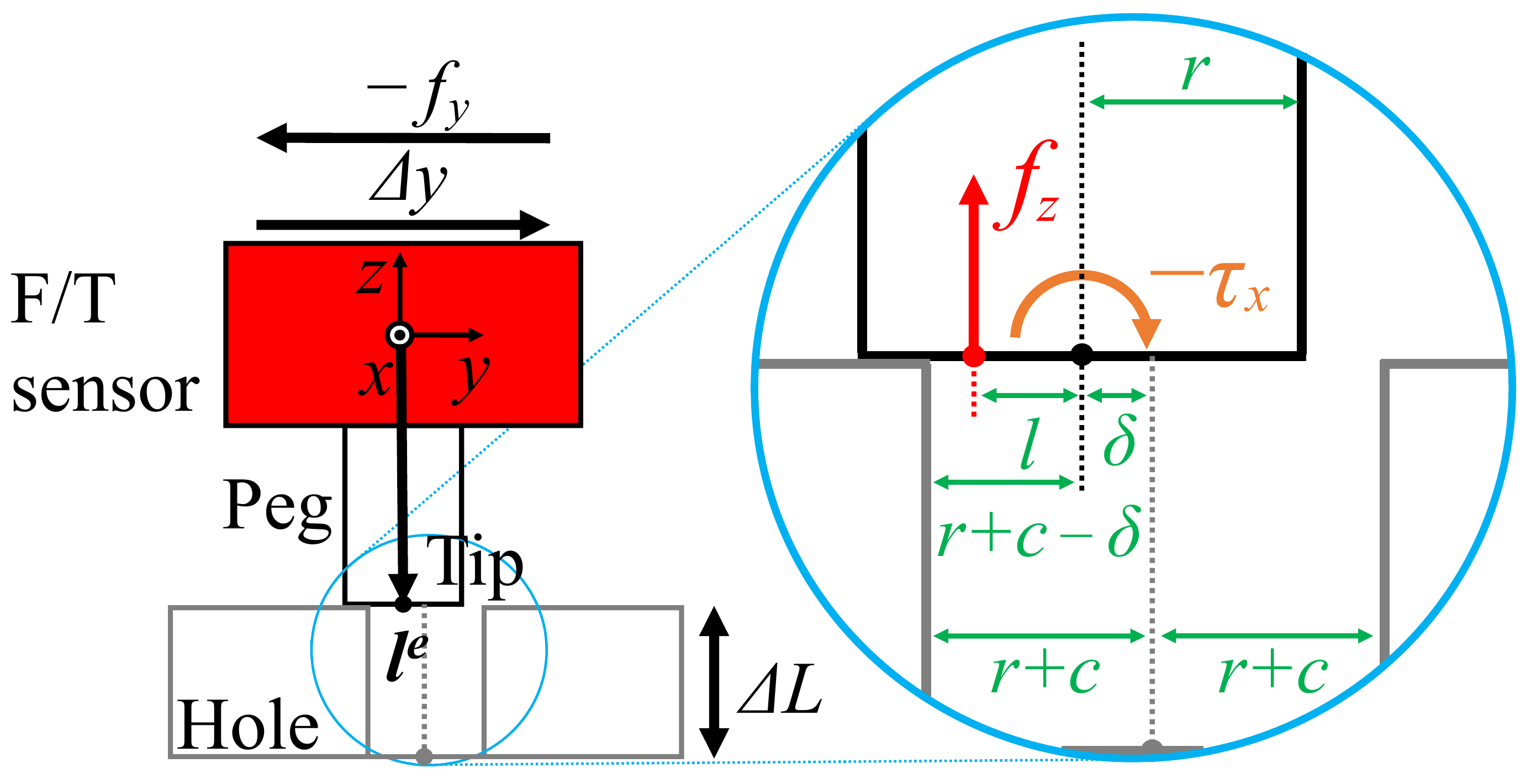}
        \end{center}
        \caption{Peg-in-hole (interference in +y-axis). 
        Variables: radius of peg ($r$), clearance ($c$), 
        distance between centers of hole and peg ($\delta$), 
        depth of position command ($\Delta L$), friction coefficient ($\mu$), 
        and distance beween the center of peg and the center of pressure ($l$) }
        \label{peg-in-hole_+y}
\end{figure}

In this subsection, we describe a method for designing non-diagonal elements 
that conduct a PiH task as one typical example. 
Based on the discussions in the previous subsections, a triangular matrix is adopted 
because it expands the range over which all eigenvalues are positive and simplifies the design.
%The PiH is a benchmark task in assembly 
%that simplifies insertion phase found in general assembly. As described in Section \ref{Asymmetric matrix}, when there are two sets of 
%interfering axes, the non-diagonal elements in an asymmetrically arranged stiffness matrix can assume a wider range 
%of values than those in a symmetrically arranged stiffness matrix. Furthermore, it is possible to design only axes 
%to be interfered. Therefore, the stiffness matrix is designed to be asymmetrical according to the PiH task. 
%Here, Table 4.1 shows each parameter from the PiH task. 
Fig. \ref{peg-in-hole_+y}
%, \ref{peg-in-hole_-y}, \ref{peg-in-hole_+x}, and \ref{peg-in-hole_-x} 
shows a schematic of the PiH task in a three-dimensional space during the search. 
This figure shows that the contact position changes depending on the displacement $\Delta y$ 
and that the sign of the moment $\tau_x$ should also change accordingly. 
Then, an effective design on a non-diagonal element $k_{yr_x}$ would induce displacements $\Delta y$. 
In the same token, $k_{xr_y}$ would also induce displacements $\Delta x$. 
%effective design would induce displacements $\Delta x$ and $\Delta y$ and the moments $my$ and $mx$ that had mutually opposed signs (positive and negative). 
%which are non-diagonal elements that cause the interference, is shown. 
To design $k_{xr_y}$ and $k_{yr_x}$, first, we consider the contact position of Fig. \ref{peg-in-hole_+y}. 
Since $c\ll|\delta|$, we set $|\delta| - c \fallingdotseq |\delta|$. 
The amount of interference by the non-diagonal elements is expressed as follows:
\begin{eqnarray}
  \delta &=& k^{\mathrm{inv}}_{yy} f_{y} + k^{\mathrm{inv}}_{yr_x} \tau_{x} \label{eqnation2_+y}
\end{eqnarray}
Here, $k^{\mathrm{inv}}_{yy}$ is considered, in addition to $k^{\mathrm{inv}}_{yr_x}$, because the external 
force on y-axis directly interferes with the movement in y-axis. 
Conversely, other non-diagonal elements are not considered 
because they do not have a direct relationship to the movement in the y-axis. 

Substituting $f_{y} = -{\mu} f_z$ and $\tau_{z} = -f_zl$ to (\ref{eqnation2_+y}), 
\begin{eqnarray}
  \delta &=& -\mu k^{\mathrm{inv}}_{yy} f_{z} + k^{\mathrm{inv}}_{yr_x} l f_z \label{eqnation2_+y}
\end{eqnarray}
Here, the position command is set $\Delta L$ below the top surface to obtain a constant 
contact force $f_z$ during the search phase. Assuming $f_z = \Delta L$ $k^{inv}_{zz}$, 
the following is obtained:
\begin{eqnarray}
  \delta &=& -\frac{(\mu k^{\mathrm{inv}}_{yy}+k^{\mathrm{inv}}_{yr_x})\Delta L}{k^{inv}_{zz}}  
  \label{eqnation2_+y}
\end{eqnarray}
where $k^{inv}_{yr_x}$, which is a non-diagonal element that is to be obtained, can be expressed as follows
by substituting $\mu = 0$ into (\ref{eqnation2_+y}). In the same token, $k_{xr_y}$ can also be derived:
\begin{eqnarray}
  k^{\mathrm{inv}}_{yr_x} &=& k^{\mathrm{inv}}_{xr_y} = -\frac{\delta_{set} k^{inv}_{zz}}{l_{set}\Delta L}
  \label{equation_kinvyrx}
\end{eqnarray}
Here, $\delta_{set}$ denotes the maximum possible error and 
$l_{set}$ denotes the distance to the center of pressure when the maximum possible error occurs.  
%\begin{eqnarray}
%   &=& -\fric{\delta k^{inv}_{zz}}{l\Delta L}
%\end{eqnarray}

%$\cdots$ \\
%and the following can be expressed when the friction coefficient $\mu = 0$:
%$\cdots$ \\
%When the friction coefficient $\mu = 0$,
%$\cdots$ \\
%and since $k_{rxy}$ = 0, the following can be expressed:
%$\cdots$ \\
%Therefore, the non-diagonal element $k_{yrx}$ can be expressed as follows:
%$\cdots$ \\
%and when the friction coefficient $\mu = 0$, it can be expressed as follows:
%$\cdots$  (62)\\
%\textcolor{red}{
%This stiffness matrix can be used when $l > 0$. Similarly, the contact positions in Fig. \ref{peg-in-hole_-y} can be expressed 
%in the same way as in Eq. (62). When the non-diagonal elements kxry in Figs.  \ref{peg-in-hole_+x} and \ref{peg-in-hole_-x} are calculated in 
%the same way as in Eqs. (50)–(62), then the follow equation is obtained:
%$\cdots$ (63)\\
%}
%
%These designed non-diagonal elements, enable the system to automatically guide the peg to the center of 
%the hole, as shown in Figs. \ref{peg-in-hole_+y}–\ref{peg-in-hole_-x}. The stiffness matrix during the Search phase is obtained by Eqs. (62) and (63). 

By setting up the non-diagonal element using (\ref{equation_kinvyrx}), 
the deviation $\delta_{set}$ is generated when the peg is not inserted, despite the command 
beneath the top surface. 
%the maximum deviation $\delta$ is generated when the peg is inserted at the assumed length $\Delta L$. 
%However, the same stiffness matrix cannot be used during the Insertion phase because 
%the contact model drastically changes from search phase to insertion phase.  
The parameters of the stiffness matrix should be changed depending on the task 
and contact state during the task. 
%This instability could extend to interactions with the environment, even if the stiffness matrix has been designed 
%as a passive element. Thus, a change in the contact model poses the risk of instability, which can be attributed 
%to the effect of the interference term that occurs when acting on a point other than the compliance center. 
Since the tip of the peg is set as the compliance center in this study, errors may occur when the contact point is away from the compliance center, depending on the interference of translational forces 
$f_x$ and $f_y$ and moments $\tau_x$ and $\tau_y$.

\begin{figure}[tb]
    \begin{minipage}{0.48\hsize}
        \begin{center}
            \includegraphics[width=40mm]{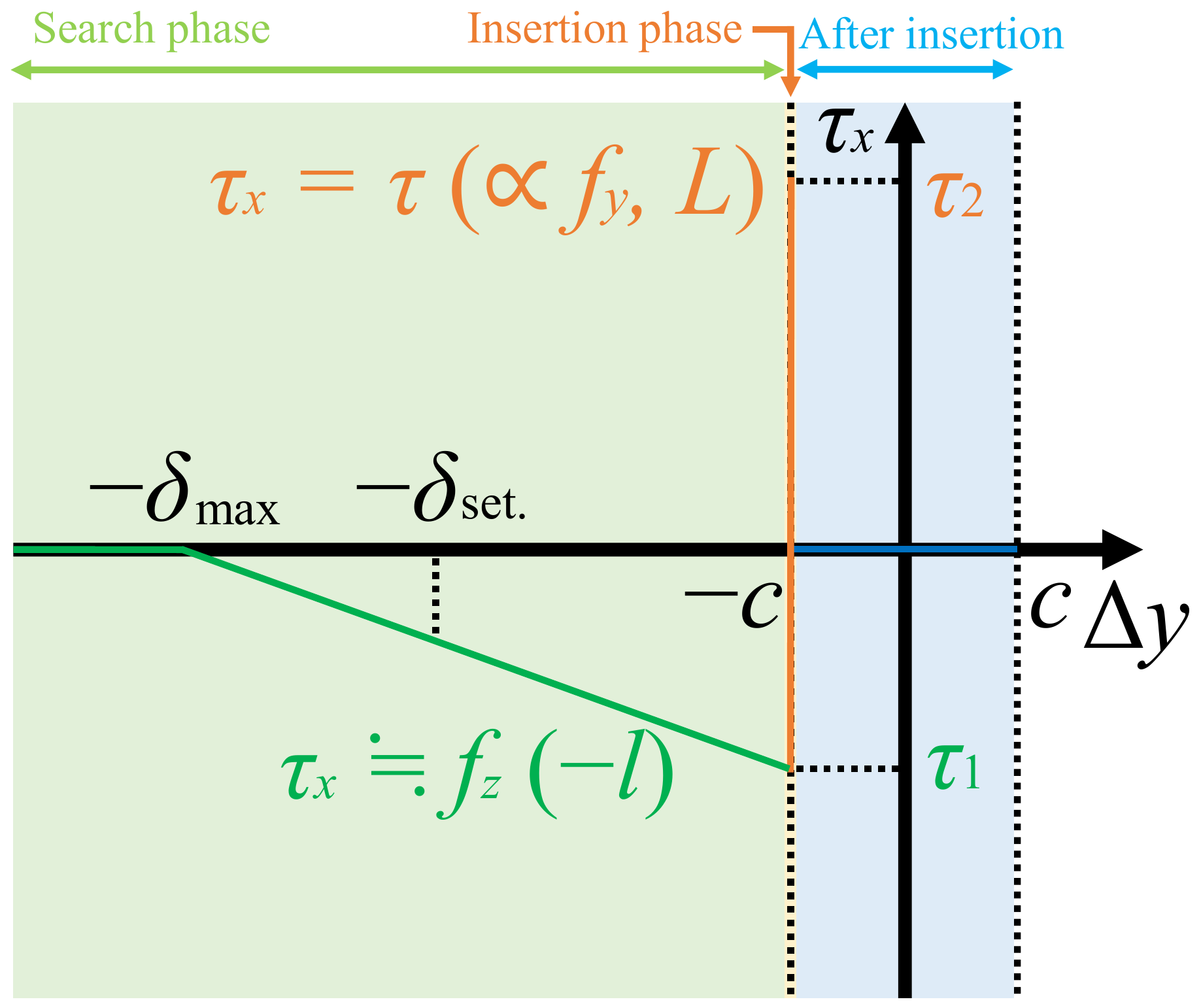}
        \end{center}
        \caption{Relation between $\tau_x$ and $\delta_x$ in negative region}
        \label{peg-in-hole_moment_-d}
    \end{minipage}
    \begin{minipage}{0.48\hsize}
        \begin{center}
            \includegraphics[width=40mm]{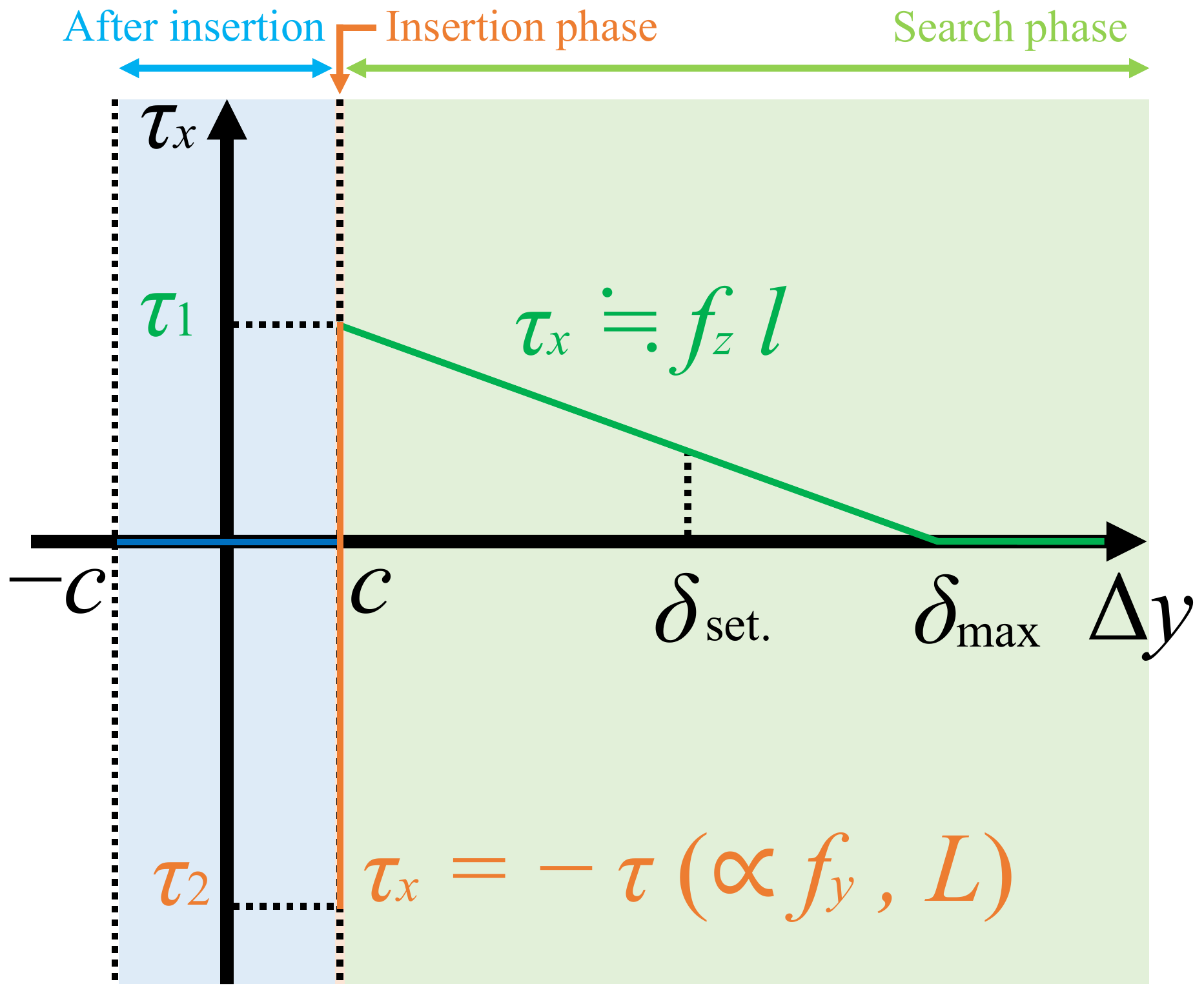}
        \end{center}
        \caption{Relation between $\tau_x$ and $\delta_x$ in positive region}
        \label{peg-in-hole_moment_+d}
    \end{minipage}
\end{figure}

\begin{figure}[tb]
    \centering
        \includegraphics[width=8.0cm]{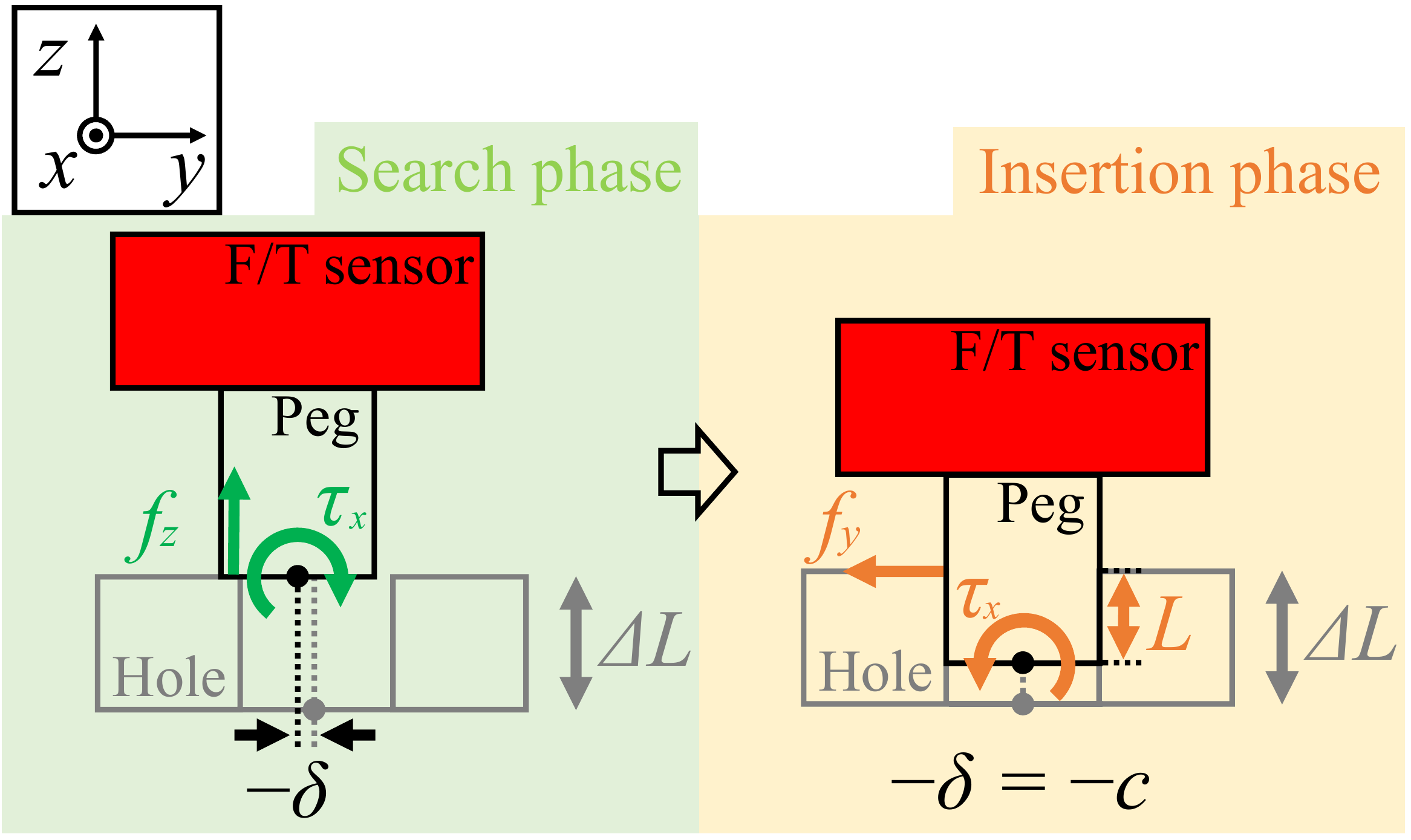}
        \caption{Force and torque during search and insertion phases}
        \label{peg-in-hole_moment_-d_detail}
\end{figure}

The contact position and force in each axis fluctuate discontinuously as the contact states change. 
The errors described above also occur in proportion to the magnitude of the contact position variation.  
Even if the compliance control is designed to be passive, hunting occurs due to these factors. 
As mentioned in the previous section, there is no theoretical restriction on the non-diagonal elements 
that make the eigenvalues of the triangular matrix positive. 
However, to avoid hunting, this error must be assumed, and this gives a substantial constraint in the 
setting of the non-diagonal elements. 
If the maximum force error expected for the task is estimated, substituting the maximum force 
into (\ref{eqnation2_+y}) yields the maximum width of the hunting movement. 
It is necessary to set the non-diagonal elements so that the hunting width is within acceptable limits. 

\if 0
The tip of the peg needs to be used as the reference point of the force sensor 
coordinate system. Fig. \ref{hosei} shows the length $\bm{l}_e$ from the detection position of the force sensor to the tip 
of the peg. This includes the length of the peg, that of the jig that fixes the peg to the sensor, and that 
from the top of the jig to the tip of the force sensor. Here, the jig and peg stiffness are set to be high 
to ensure that they are firmly fixed to the force sensor. There are no issues with the translational forces 
$f_x$, $f_y$, and $f_z$ or the moment mz even if the force sensor coordinate system changes. However, since the 
moments mx and my have different values at the force sensor detection position and at the tip of the peg, 
there is a need to calculate the moment applied to the tip of the peg from the force sensor detection 
position and convert it. Setting the translational force and moment at the force sensor detection position 
as fsensor and msensor, then the moment $\bm{\tau}^{\mathrm peg}$mpeg at the tip of the peg is expressed as follows:
%\begin{eqnarray}
%    \bm{\tau}^{\mathrm {peg}} &=&  \bm{\tau}^{\mathrm {sensor}} - \bm{l}^{\mathrm {e}}\bm{f}^{\mathrm {sensor}}\label{m_peg}
%  \end{eqnarray}
  
\begin{figure}[ht]
  \centering
      \includegraphics[clip,width=5.0cm]{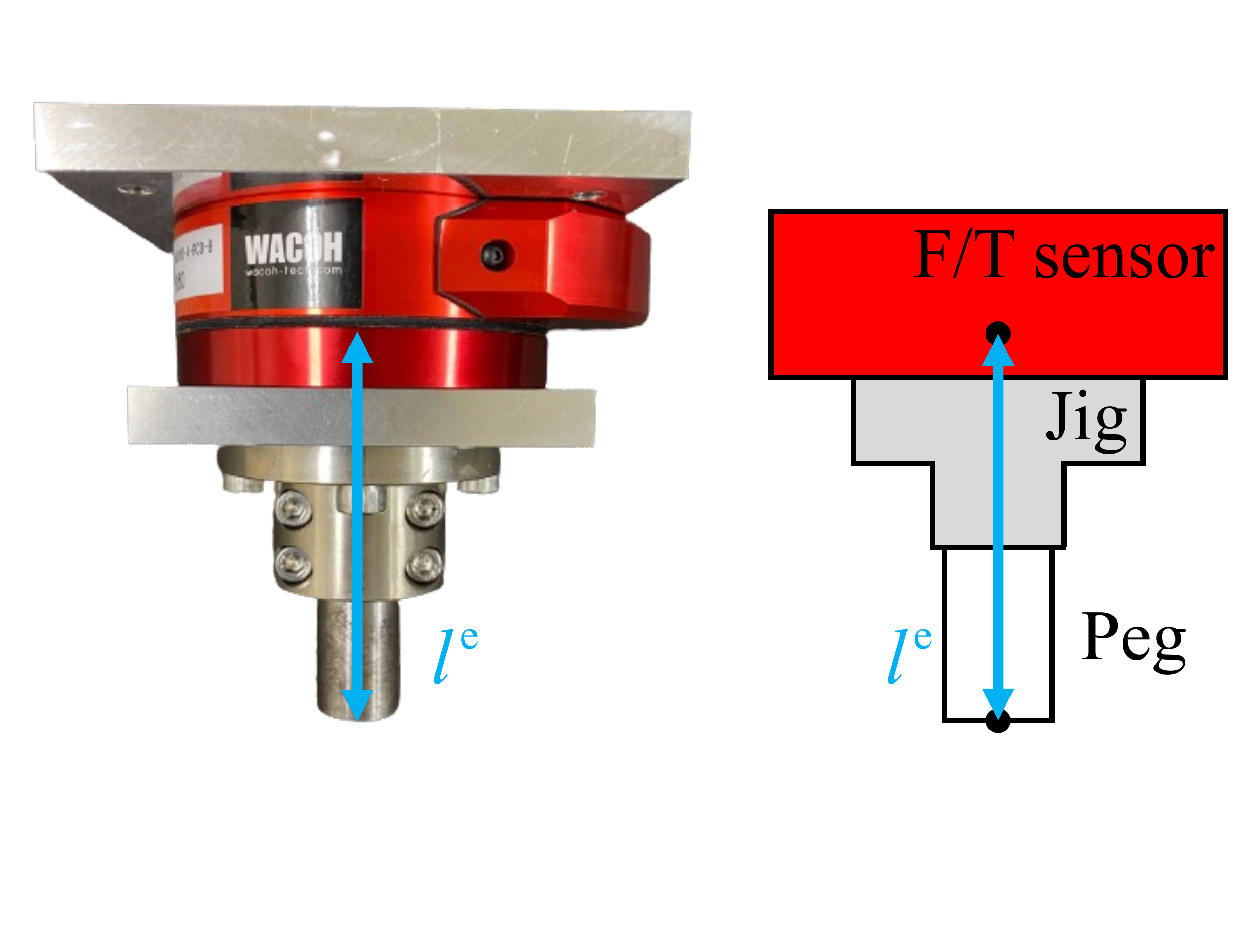}
      \label{hosei}
\end{figure}
\fi 
\section{EXPERIMENT}
\subsection{Experimental device setup}
\begin{figure}[h]
    \begin{minipage}{0.48\hsize}
        \begin{center}
            \includegraphics[width=30mm]{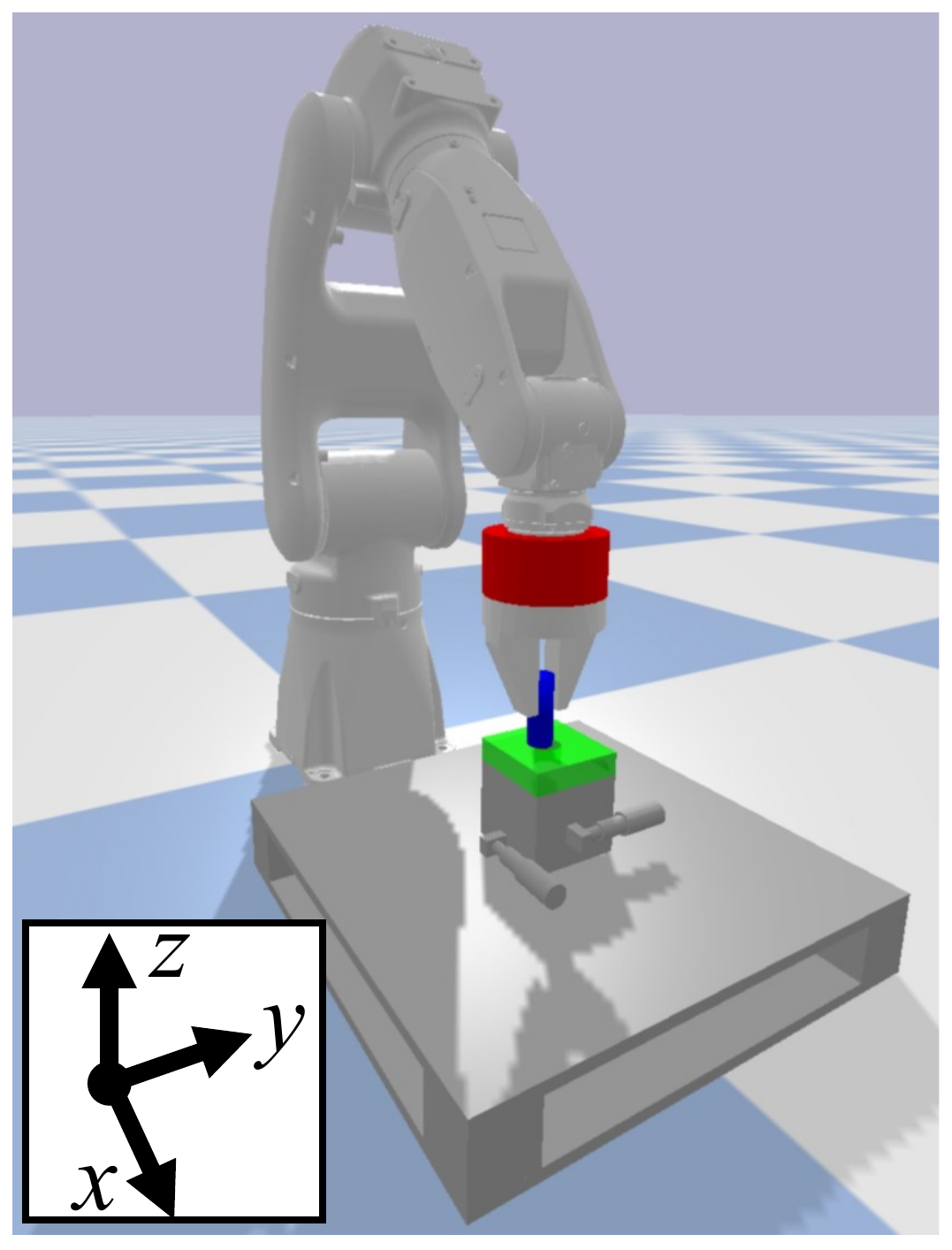}
        \end{center}
        \caption{Simulation setup}
        \label{robot_simulation}
    \end{minipage}
    \begin{minipage}{0.48\hsize}
        \begin{center}
            \includegraphics[width=30mm]{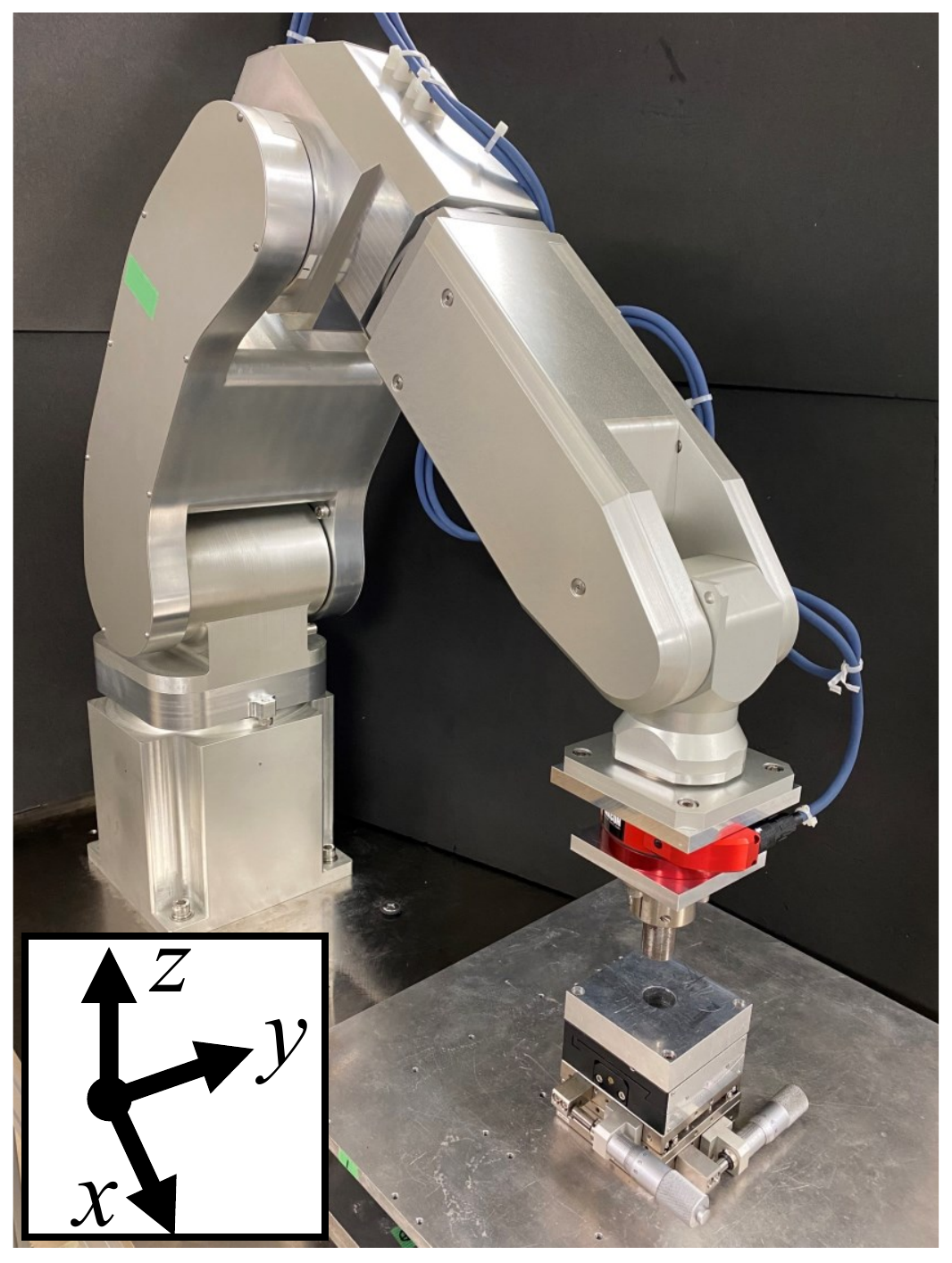}
        \end{center}
        \caption{Experimental setup}
        \label{robot_real}
    \end{minipage}
\end{figure}

We conducted experimental evaluations using a real robot. Simulations based on the 
parameters of the actual machine were also used to evaluate quantitative comparisons 
from a safety perspective. Both used an industrial six-axis manipulator, and a six-axis 
force sensor was attached to the wrist of the robot. 
Fig.~\ref{robot_simulation} shows the manipulator used in the simulation, 
the coordinate system, and the manipulator used in the real robot. 
Table~\ref{table:1} presents  the specifications of the manipulator used in the real robot. 
The manipulator used in the simulation had the same parameters as the experimental device. 
%Table~\ref{table:2} presents the specifications of the six-axis force sensor. 
%Fig. ?  shows the block diagram of the control used in the simulation and the real 
%robot experiment, 
%and 
Table~\ref{table:3} lists each control parameter. 

\begin{table}[tb]
	\centering
	\caption{Specifications of manipulator}%タイトル
	\label{table:1}
	\begin{tabular}{r|r|r|r}
		\hline
		 & Max. torque & Max. velocity & Link length from \\			
		 & [Nm] & [rad/s] & previous joint [mm]\\			
		\hline
		Joint 1 & 68.5 & 1.70 &  200 \\
		Joint 2 & 73.0 & 1.59 &  130 \\
		Joint 3 & 27.5 & 2.12 &  345 \\
		Joint 4 & 12.9 & 2.26 &  100 \\
		Joint 5 & 11.0 & 2.65 &  240 \\
		Joint 6 & 7.4  & 4.56 &  80 \\
		\hline
	\end{tabular}
\end{table}
\begin{table}[tb]
	\centering
	\caption{Control parameters}%タイトル
	\label{table:3}
	\begin{tabular}{l|c|l}
		\hline
		Proportional gain &$\bm{K}_p$& diag[600, 600, 600, 600, 600, 600] \\
		Derivative gain   &$\bm{K}_d$& diag[50.0, 50.0, 50.0, 50.0, 50.0, 50.0]\\
		Moment of inertia & $\bm{I}$ & diag[2.0, 1.0, 0.8, 0.25, 0.15, 0.05]\\
		%[kgm$^2$] & & \\
		~[kgm$^2]$ & & \\
		Mass of admittance & $\bm{M}$ & diag[1.0, 1.0, 1.0, 1.0, 1.0, 1.0] \\
		model [kg] & & \\
		Damper of admittance  & $\bm{D}$ & diag[50.0, 50.0, 50.0, 5.0, 5.0, 5.0] \\
		model [Ns/m] & & \\
		%Stiffness of admittance model [N/m] & $\bm{K}$ &\\
		Cutoff frequency of & $g$ & 1.5 \\
		derivative filter [Hz] & & \\
		Samping time [s] & $T_s$& 0.001\\
		\hline
	\end{tabular}
\end{table}

\subsection{Simulation stability comparison}

First, we evaluated the performance of the proposed method by simulating the induction by 
stiffness ellipses without limitation of maximum external force. 
%Since previous studies have shown that a symmetric matrix has 
%an ability to induce motion, 
We compared the inductive effect of the proposed method with an 
asymmetric matrix to that of a symmetric matrix. A vertical command trajectory on the z-axis was 
given with the maximum command depth $\Delta L = 20$[mm]. Then the robot with a 20 mm diameter 
peg made contact with the top surface of the plate and the motion was induced in the y-axis after contact.  
Here, the friction coefficient $\mu$ was 0.4. 
 
\begin{figure}[b]
        \centering
        \includegraphics[clip,width=8.4cm]{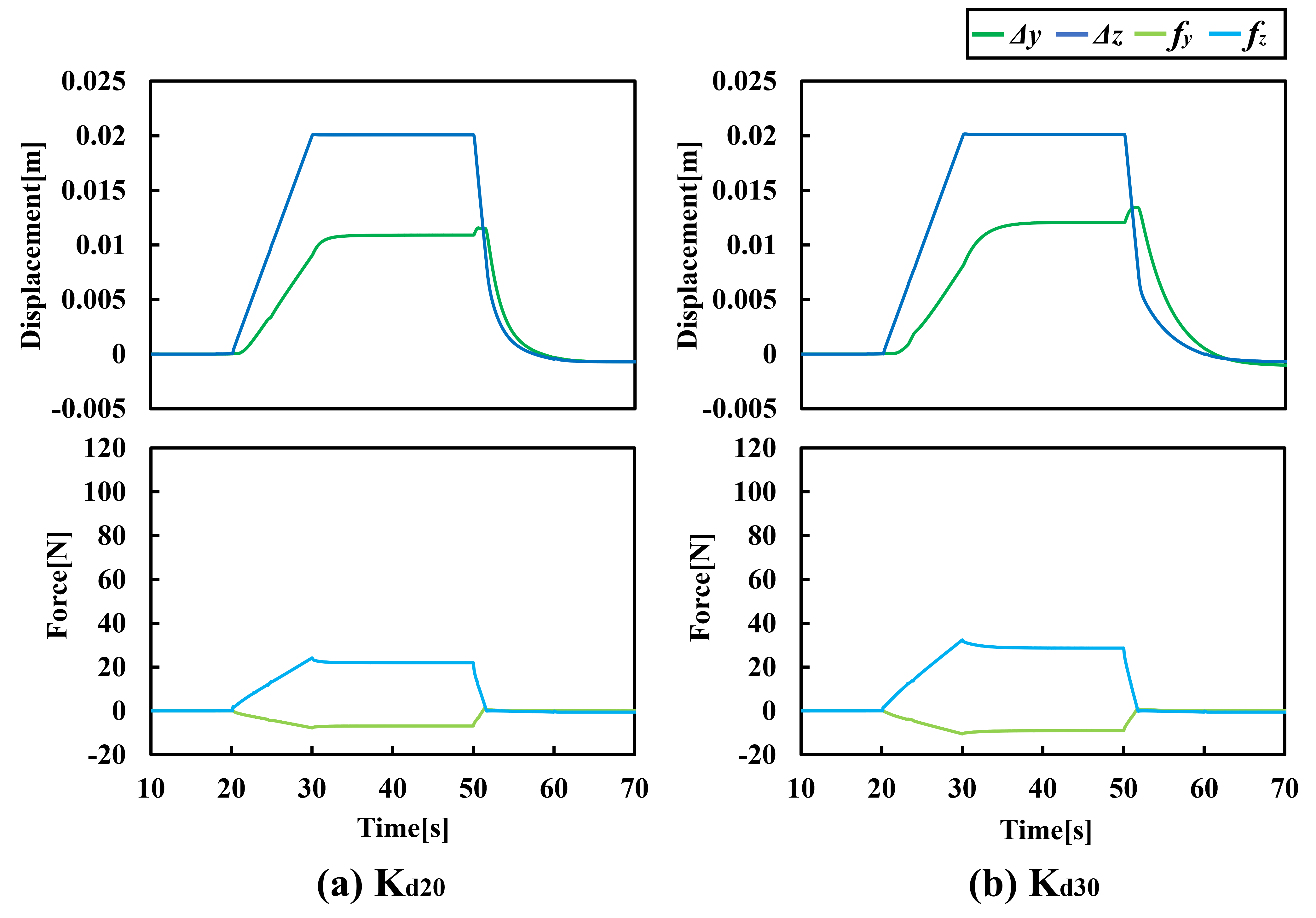}
        \caption{Responses with each stiffness matrix with different interference design. (symmetric) }
        \label{result_sim2_1}
\end{figure}

Fig.~\ref{result_sim2_1} shows the simulation results using symmetric matrices. 
Here, $\bm{K}_{d20}$ and $\bm{K}_{d30}$ denote the symmetric matrices designed 
to induce the motion in the y-axis with 20 mm and 30 mm lengths, respectively. 
The results in Fig.~\ref{result_sim2_1} show that the motion was induced in the desired direction 
with the use of $\bm{K}_{d20}$, but no significant increase in the response values 
was observed with the use of $\bm{K}_{d30}$. 
This is because the inclination of the stiffness 
ellipse decreases as the amount of interference increases, and  
the force acts along the direction close to the long-axis direction of 
the inclined stiffness ellipse due to the frictional force. 
In the stiffness matrices designed for induction above 30 mm, the 
inductive effect in the y-axis no longer occurred due to this effect. 

\begin{figure}[tb]
        \centering
        \includegraphics[clip,width=8.4cm]{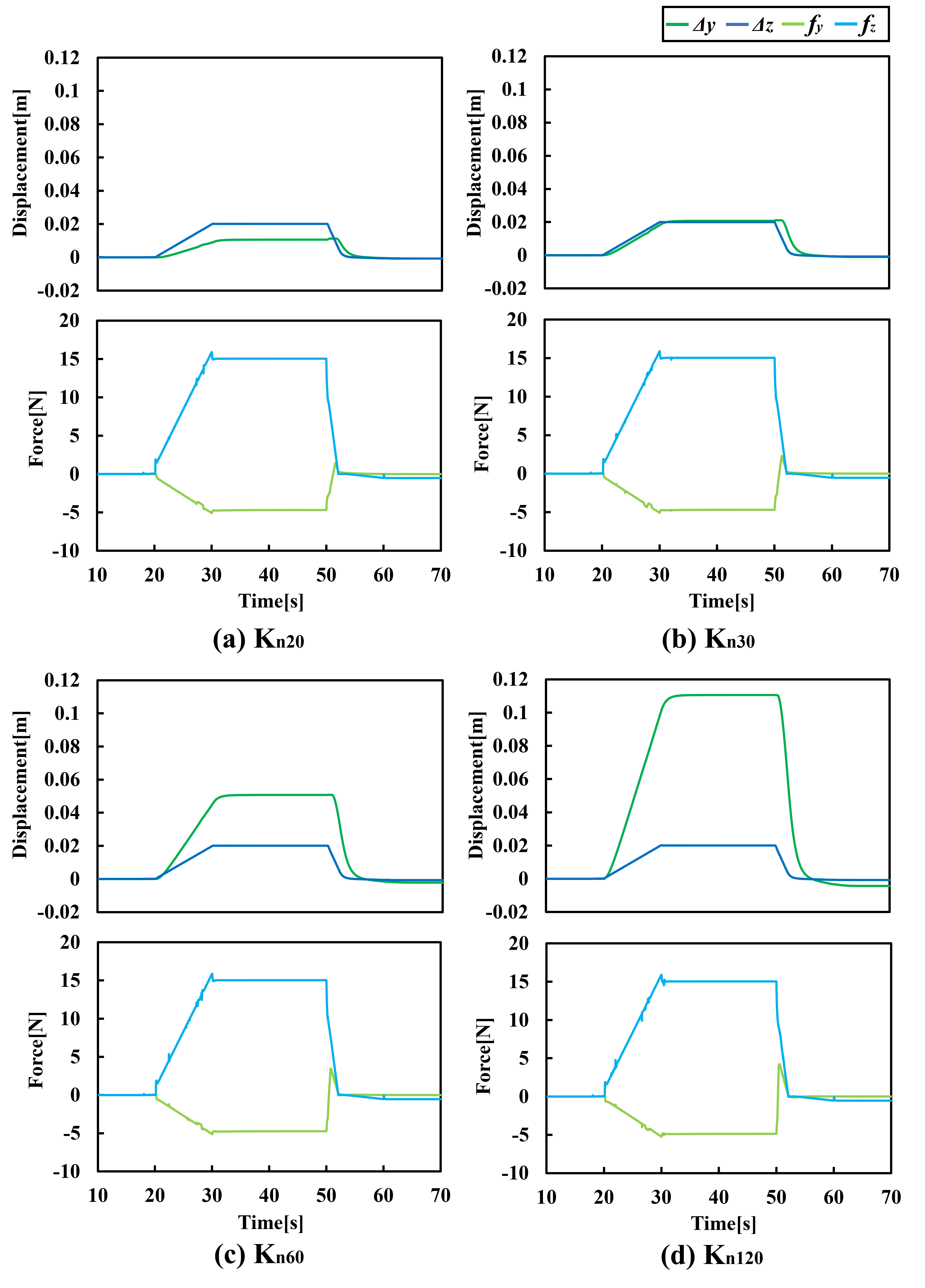}
        \caption{Responses with each stiffness matrix with different interference design. (asymmetric)}
        \label{result_sim2_2}
\end{figure}

Meanwhile, the results in Fig.~\ref{result_sim2_2} show that the interference in 
the response values were generally in proportion to the design values of the four 
interferences (20, 30, 60, and 120 mm). 
Here, the stiffness matrices were designed as follows: 
\begin{eqnarray}
  \bm{K}_{n20} \!\!\!\!&=& \!\!\!\!\left[ \begin{array}{cc}\!\! 500 & -500 \!\! \\ \!\! 0 & 750 \!\!\end{array} \right], 
  \bm{K}_{n30} = \left[ \begin{array}{cc} \!\!500 & -750 \!\! \\ \!\! 0 & 750 \!\! \end{array} \right] \nonumber\\
  \bm{K}_{n60} \!\!\!\!&=& \!\!\!\!\left[ \begin{array}{cc} \!\! 500 & -1500 \!\! \\ \!\! 0 & 750 \!\! \end{array} \right],  
  \bm{K}_{n120} = \left[ \begin{array}{cc}  \!\! 500 & -3000 \!\! \\ \!\! 0 & 750 \!\! \end{array} \right] \nonumber
\end{eqnarray}

In all cases, the response values 
were slightly below the design values due to friction in the sliding direction. This is a reasonable result 
because the robot reproduces the desired impedance and the same phenomenon occurs with RCC, 
which sets the impedance mechanically. 


As shown in Fig. \ref{result_sim2_2}, the asymmetrical arrangement induces 
interference in the y-axis at the desired force of $f_z = 15$ N. The 
amount of interference depends on the friction coefficient of the member, 
regardless of whether it is symmetrical or asymmetrical. The 
asymmetrical arrangement enables contact with the desired force; furthermore, 
the non-diagonal elements can assume a wider range of values when compared with 
the symmetrical arrangement. Therefore, in asymmetrically arranged matrices, 
the value of interference with other axes is less constrained than in 
symmetrically arranged matrices.

\subsection{Experimental result}
\begin{figure}[tb]
    \centering
        \includegraphics[width=8.4cm]{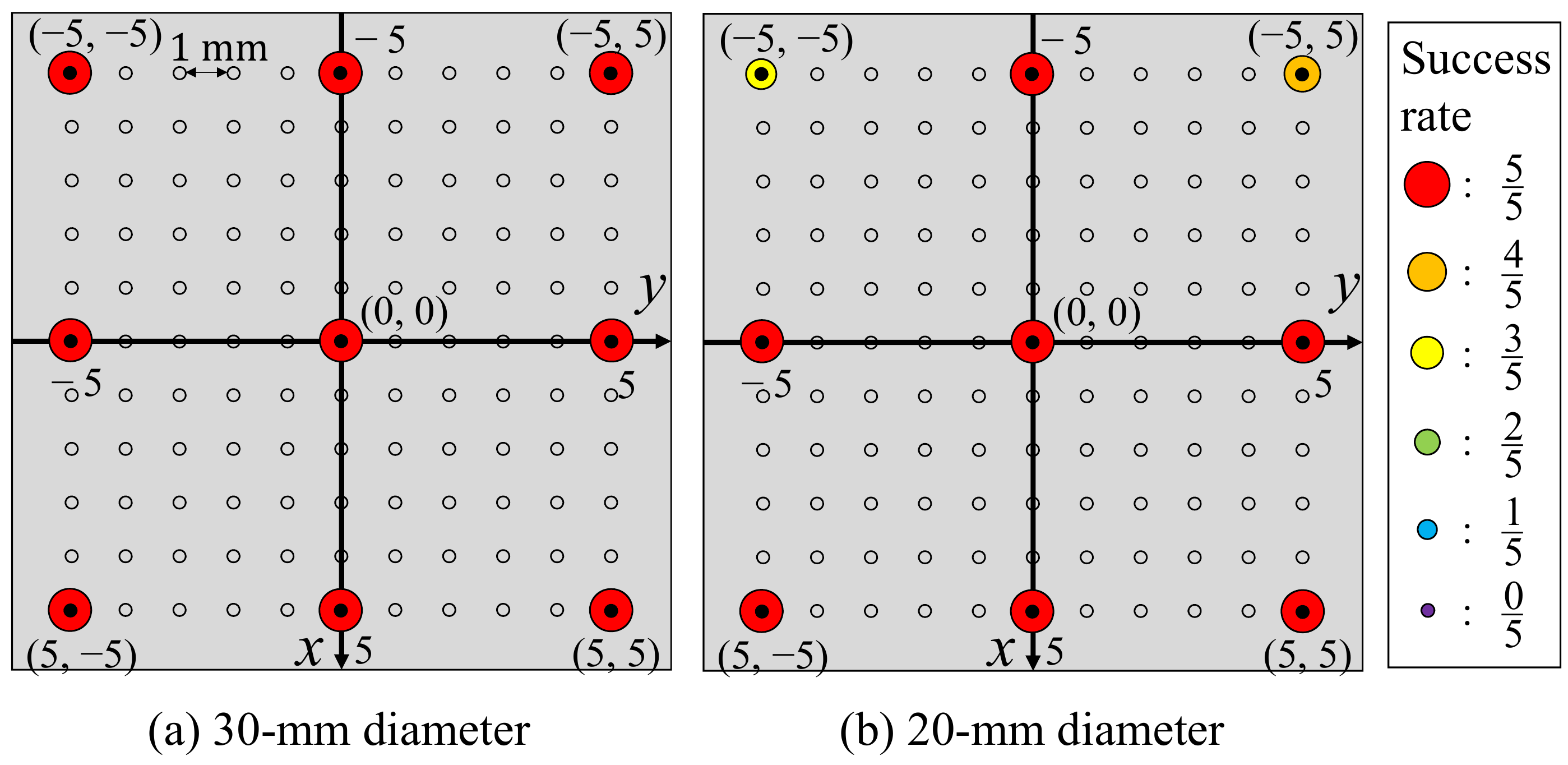}
        \caption{Success rate of PiH experiment}
        \label{fig:exp_sr}
\end{figure}
\begin{figure}[tb]
    \centering
        \includegraphics[width=8.4cm]{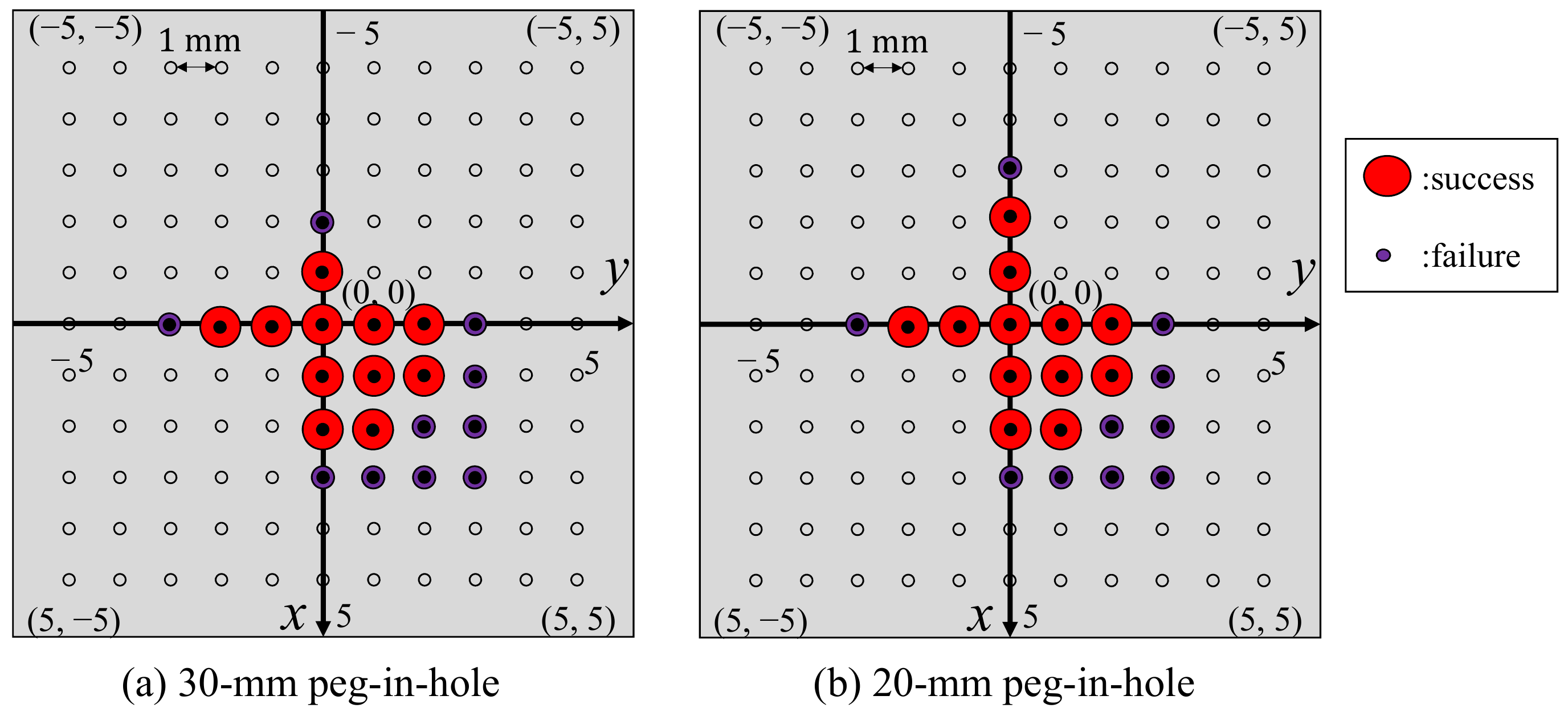}
        \caption{PiH experiment of the baseline method}
        \label{fig:exp_bl}
\end{figure}

The stiffness matrix was designed as follows:
\begin{eqnarray}
\bm{K}=\left[
\begin{array}{cccccc}
500 & 0 & 0 & 0 & k_{xr_y} & 0 \\
0 & 500 & 0 & k_{yr_x} & 0 & 0 \\
0 & 0 & 500 & 0 & 0 & 0 \\
0 & 0 & 0 & 50 & 0 & 0 \\
0 & 0 & 0 & 0 & 50 & 0 \\
0 & 0 & 0 & 0 & 0 & 50 
\end{array}
\right]
\end{eqnarray}
Assuming that an error of $\pm$5 mm would be generated in each direction of 
the x- and y-axes, we designed $k_{xr_y}$ and $k_{yr_x}$ using (\ref{equation_kinvyrx}), 
and conducted PiH with a 30-mm diameter (clearance of 420 $\mu$m) and 20-mm diameter 
(clearance of 40 $\mu$m) peg. Fig.~\ref{fig:exp_sr} shows the number of successes 
and failures for different positional errors. The experiments show that the 
proposed method has the ability to perform PiH via adjustment of the stiffness matrix and without 
environment-adaptive trajectory planning. 
RCC, which also performs PiH with only stiffness adjustment, was employed as a baseline 
method. In this experiment, however, the behavior was reproduced, not by the mechanism, 
but by the control.
For the 30-mm diameter (420 $\mu$m clearance), all locations resulted in successful 
insertion for five out of five attempts; and for the 20-mm diameter (40 $\mu$m clearance), 
all but two locations resulted in task completion. 
%In this way, PiH with a clearance of 420 $\mu$m and 40 $\mu$m, respectively, was achieved. 

The results of implementing the same design as the RCC in an admittance control are 
shown in Fig.~\ref{fig:exp_bl}. 
The success rates were evaluated over a sufficiently long time for insertion to 
ensure that implementation by control, rather than mechanism, would not lower the 
evaluation. 
In general, RCCs are designed to passively displace only its position, without changing its 
posture, in response to external forces in the insertion direction.
Therefore, the position error of the peg is compensated by the contact force from 
the edge of the hole acting in the direction of the center of the hole.
However, if the error is large, the external force from the edge of the hole will 
not guide the peg to the direction of the hole, and the error correction range 
will be limited. 
In contrast, the proposed method designs the non-diagonal component to guide 
the peg in the direction of the hole according to the moment. Therefore, the position 
error was compensated for even when the error was larger. 
In other words, it was confirmed that the error tolerance of PiH can be expanded 
by extending the setting range of the stiffness matrix to a region that has not 
been considered in the past.

\begin{figure}[tb]
    \centering
        \includegraphics[width=8.0cm]{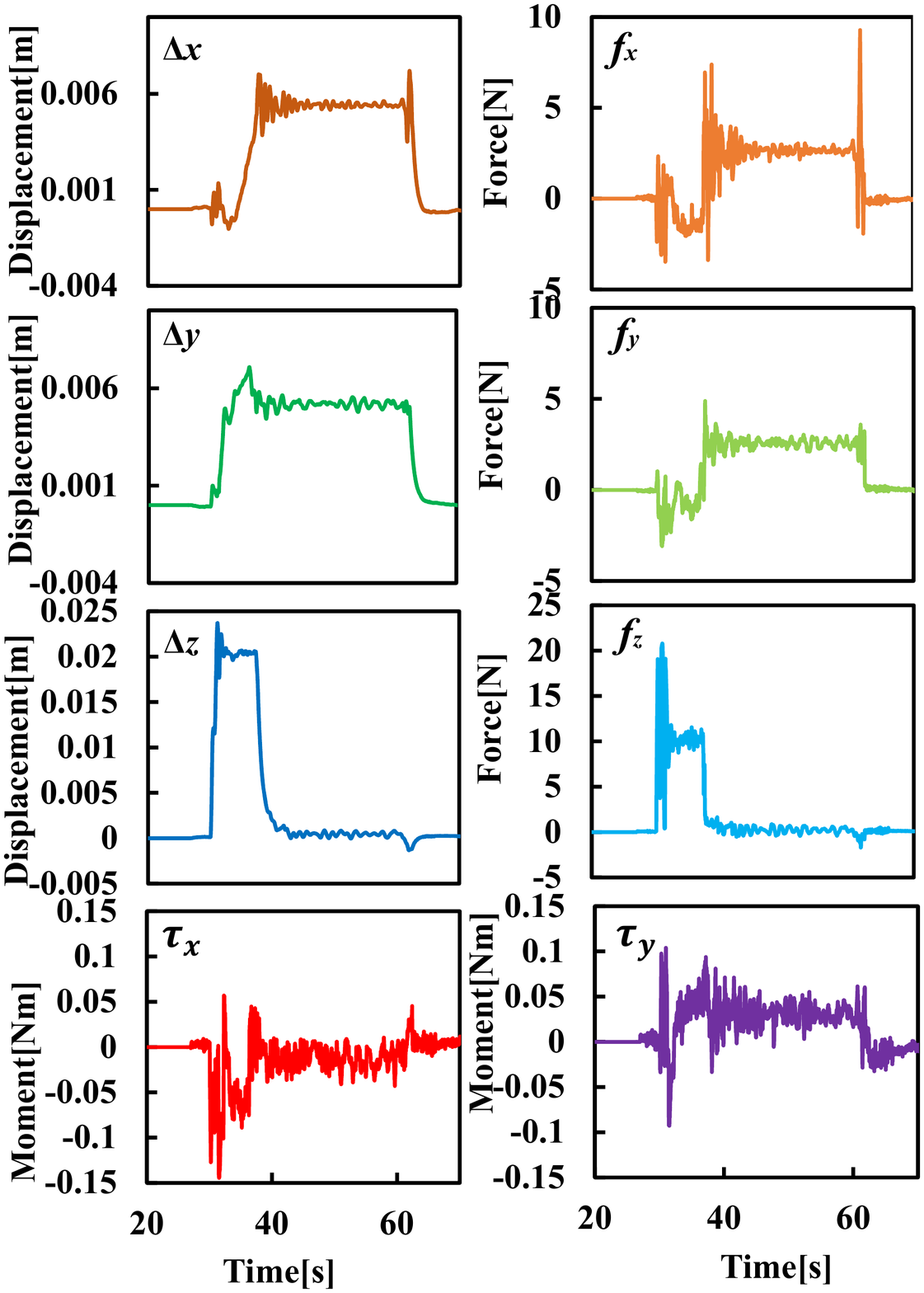}
        \caption{Responses during insertion of 20-mm diameter peg}
        \label{exp_res}
\end{figure}

Next, the positional displacement command and force value 
determined at each axis by the stiffness matrix at an error of 
$\Delta(x, y) = (-5, -5)$ for a 20-mm diameter PiH are shown in 
Fig.~\ref{exp_res}. It can be seen from the results around time 30 [s] that contact with the hole 
edge results in the application of a force $f_z$ in the insertion direction. 
This generates $\tau_y$ and $\tau_x$ according to its magnitude. Finally, these 
moments interfere with $\Delta y$ and $\Delta x$, respectively, due to 
the non-diagonal elements. Compared with the PiH of the 30-mm case, the 
20-mm diameter peg had a narrower clearance and an environment where the 
moment was less likely to occur. For this reason, there was one failure out of 
five for the $\Delta(x, y) = (-5, 5)$ case and two failures out of five for 
the $\Delta(x, y) = (-5, -5)$ case. 
Nevertheless, all the pegs were inserted in a stable manner in the areas 
where the error was small.

\begin{figure}[tb]
    \centering
        \includegraphics[width=8.0cm]{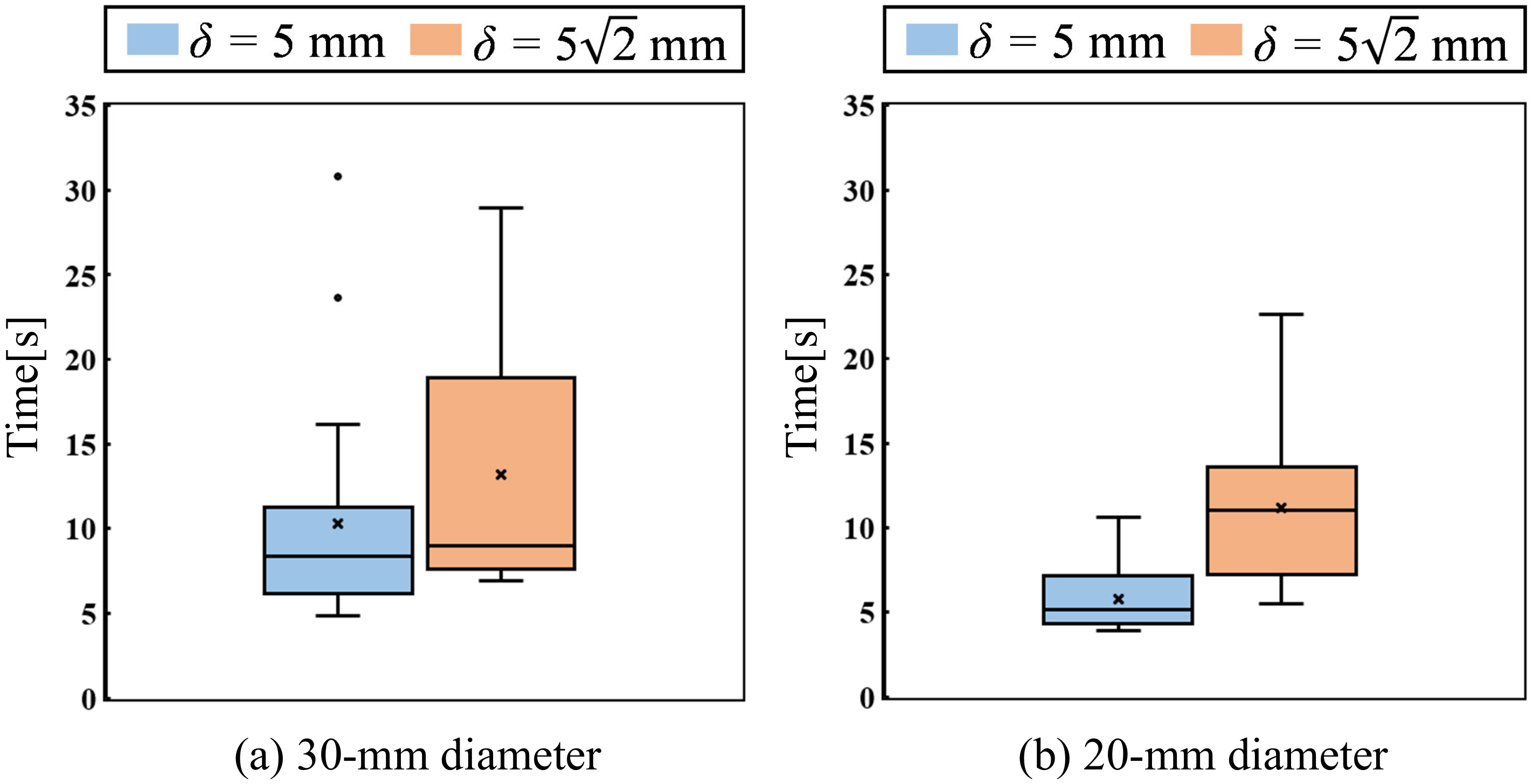}
        \caption{Insertion time of PiH experiment}
        \label{fig:exp_time}
\end{figure}

Fig.~\ref{fig:exp_time} shows the time required for insertion. 
For both the 30-mm and 20-mm diameter pegs, insertion was achieved 
with less variation and in a shorter time for an error of 5 mm compared with 
an error of $5\sqrt{2}$ mm. This is because the distance to the hole increased 
when the error increased to $5\sqrt{2}$ mm. Thus, more time was required for 
task completion. Furthermore, at the increased error level, it was difficult 
to generate a moment with the contact position due to the errors generated along 
both the x- and y-axes. In comparison, the insertion of the 20-mm diameter peg 
exhibited less variation and a shorter average time than the 30-mm diameter peg. 
In this experiment, the stiffness matrix was designed on the assumption that 
the friction coefficient was zero. Therefore, the frictional force became 
a source of error, but a larger effect was present in the case of the 30-mm 
diameter peg, which had a larger contact surface area than that of the 20-mm diameter peg.

\section{Conclusion}
In this study, we proposed a method for designing non-diagonal elements in a 
stiffness matrix. We described the extent to which non-diagonal elements that achieve 
the desired amount of interference could be designed, and the cases of symmetrical and 
asymmetrical arrangements. 
To obtain stable contact with a symmetric matrix, the matrix should be positive definite, 
i.e., all eigenvalues must be positive, however its parameter design is complicated. 
In this study, we therefore focused on the use of asymmetric matrices in compliance control  
and showed that the design of eigenvalues can be simplified by using a triangular matrix. 
This approach expands the range of the stiffness design and enhances the ability 
of the compliance control to induce motion. The experiments using the stiffness 
matrix based on the proposed method, demonstrated that assembly is possible with 
just a simple trajectory. In addition, it showed that precision assembly with a clearance of 
several dozens of microns is possible.

\end{document}